%% file: main.tex
\documentclass[10pt,twocolumn,letterpaper]{article}

\usepackage[pagenumbers]{cvpr} 


\definecolor{cvprblue}{rgb}{0.21,0.49,0.74}
\usepackage[pagebackref,breaklinks,colorlinks,allcolors=cvprblue]{hyperref}
\usepackage{multirow}
\usepackage{adjustbox}
\usepackage{algorithm}

\usepackage{xspace}
\makeatletter 
\DeclareRobustCommand\onedot{\futurelet\@let@token\@onedot} 
\def\@onedot{\ifx\@let@token.\else.\null\fi\xspace}  
\def\eg{\emph{e.g}\onedot} 
 
\def\ie{\emph{i.e}\onedot}

\def\etal{\emph{et al}\onedot} 
\makeatother 

\usepackage[T1]{fontenc}

\def\paperID{****} 
\def\confName{CVPR}
\def\confYear{2026}

\title{AVI-Edit: Audio-sync Video Instance Editing with Granularity-Aware Mask Refiner}

\author{Haojie Zheng\textsuperscript{1,2\dag}\quad
Shuchen Weng\textsuperscript{2,3\dag}\quad
Jingqi Liu\textsuperscript{1,2}\quad
Siqi Yang\textsuperscript{5}\\
Boxin Shi\textsuperscript{3,4\ddag}\quad
Xinlong Wang\textsuperscript{2}
\\
\parbox{\textwidth}{\centering \small
{\textsuperscript{1}School of Software and Microelectronics, Peking University}\quad
{\textsuperscript{2}Beijing Academy of Artificial Intelligence}\\
{\textsuperscript{3}State Key Lab of Multimedia Info. Processing, School of Computer Science, Peking University}\\
{\textsuperscript{4}Nat’l Eng. Research Ctr. of Visual Tech., School of Computer Science, Peking University}
\\
{\textsuperscript{5}Institute for Artificial Intelligence, Peking University}\\
{\tt\small \{suimu, liujingqi\}@stu.pku.edu.cn}\quad
{\tt\small \{scweng, wangxinlong\}@baai.ac.cn}\\
{\tt\small \{yousiki, shiboxin\}@pku.edu.cn}\vspace{-1mm}
}
}

\begin{document}

\input{fig/teaser}   
\let\thefootnote\relax
\footnotetext{
    \begin{minipage}[t]{\textwidth}
        $^\dag$ Equal contribution. \\
        $^\ddag$ Corresponding author.
    \end{minipage}
}

\input{sec/0_abstract}

\input{sec/1_intro}

\input{sec/2_related}

\input{fig/pipeline}

\input{sec/3_dataset}

\input{sec/4_method}

\input{sec/5_experiment}

\input{sec/6_conclusion}

\input{sec/X_suppl}

\clearpage
{
    \small
    \bibliographystyle{ieeenat_fullname}
    \bibliography{main}
}

\end{document}

%% file: fig/teaser.tex
\twocolumn[{%
\renewcommand\twocolumn[1][]{#1}%
\maketitle
\vspace{-10mm}
\begin{center}
    \centering
    \captionsetup{type=figure}
    \includegraphics[width=\linewidth]{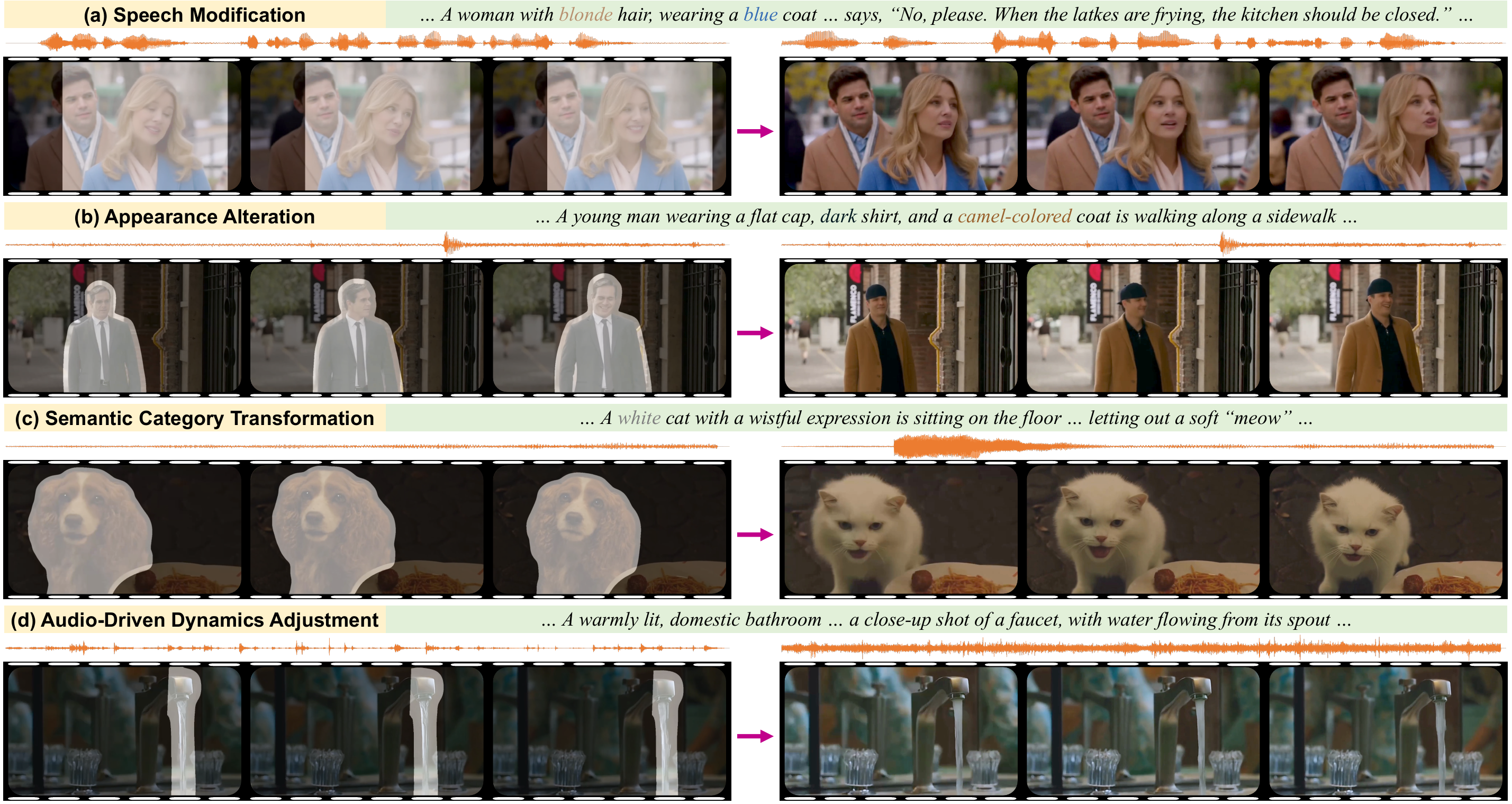}
    \vspace{-7mm}
    \captionof{figure}{AVI-Edit effectively edits audio-sync video instance based on a coarse instance mask to indicate the target instance and a text description to specify the edit direction. 
    We demonstrate this capability across four representative scenarios:
    (a) modifying the speech of the target woman while preserving her appearance;
    (b) altering the appearance of the male character while preserving his original speech;
    (c) changing the dog into a cat while transforming its vocalization; and
    (d) adjusting the dynamics of the water flow purely via audio cues.
    Generated videos and accompanying audios are available in the supplementary material.}
    \label{fig:teaser}
    \vspace{-2mm}
\end{center}
}]

%% file: sec/0_abstract.tex
\begin{abstract}
Recent advancements in video generation highlight that realistic audio-visual synchronization is crucial for engaging content creation. However, existing video editing methods largely overlook audio-visual synchronization and lack the fine-grained spatial and temporal controllability required for precise instance-level edits. 
In this paper, we propose AVI-Edit, a framework for audio-sync video instance editing. 
We propose a granularity-aware mask refiner that iteratively refines coarse user-provided masks into precise instance-level regions. 
We further design a self-feedback audio agent to curate high-quality audio guidance, providing fine-grained temporal control.
To facilitate this task, we additionally construct a large-scale dataset with instance-centric correspondence and comprehensive annotations. 
Extensive experiments demonstrate that AVI-Edit outperforms state-of-the-art methods in visual quality, condition following, and audio-visual synchronization. Project page: \href{https://hjzheng.net/projects/AVI-Edit/}{https://hjzheng.net/projects/AVI-Edit/}.
\end{abstract}

%% file: sec/1_intro.tex
\section{Introduction}
\label{sec:intro}

Video editing methods provide users with a powerful approach to content creation (\eg, modifying the appearance of a specific person). Recent commercial models (\eg, Sora-2~\cite{sora2} and Veo3~\cite{veo3}) demonstrate that realistic audio is crucial for creating engaging experiences. This advancement naturally leads to an urgent need for methods that perform instance-level edits on videos while preserving this crucial audio-visual synchronization.

However, the vast majority of existing video editing models~\cite{contextflow, vires, odiscoedit, videopainter, gencompositor} focus exclusively on the visual features, breaking the original audio-visual synchronization. 
To introduce audio cues, AvED~\cite{aved} introduces a cross-modal contrastive scheme to address the synchronization, but it mainly focuses on scene-level alignment.
In contrast, while Object-AVEdit~\cite{objavedit} achieves object-specific control, its inversion-regeneration paradigm inherently lacks temporal controllability (\eg, specifying the precise timing of an event).
While recent work~\cite{mtv} demonstrates that audio is a fine-grained temporal guidance for video generation, this approach still remains largely unexplored for audio-sync video editing.

In this paper, we propose \textbf{AVI-Edit}, a framework for \textbf{A}udio-sync \textbf{V}ideo \textbf{I}nstance \textbf{Edit}ing. 
As illustrated in \cref{fig:teaser}, given an instance mask to indicate the target instance and a text description to specify the edit direction, AVI-Edit seamlessly modifies the target instance and its accompanying audio, while preserving the original background and non-target audio components. 
This approach thereby enables a range of applications, including modifying the speech of an active speaker (\cref{fig:teaser}~(a)), altering the appearance of the person (\cref{fig:teaser}~(b)), 
changing the semantic category of target instance (\cref{fig:teaser}~(c)), or adjusting the dynamics of the subject purely with audio cues (\cref{fig:teaser}~(d)).

We build our framework upon the pretrained Wan2.2-5B~\cite{wan} to leverage its strong generative priors in generating temporally consistent and visually appealing videos. 
To achieve precise spatial control at the instance level, we propose the granularity-aware mask refiner with the similar diffusion transformer architecture of the video backbone, which receives a precision factor and references video and audio tokens to estimate precise edited regions.
We further design the self-feedback audio agent to run a separate-generate-remix-rework cycle, which finally produces the curated audio tokens to provide explicit temporal guidance and control the event timing of edited videos.

To facilitate model training and evaluation, we further present {\sc AVISet}, split into 71k training, 1k validating, and 1k testing videos with accompanying audios. All clips are carefully filtered to ensure each contains one primary and non-silent sounding instance. We annotate all splits with instance masks and scene-level text descriptions, while the testing set additionally includes paired original–edited text instructions for fair evaluation.

Our contributions are summarized as follows:
\begin{itemize}
    \item We propose an audio-sync video instance editing framework with fine-grained spatial and temporal control, along with a tailored dataset for training and evaluation.
    
    \item We propose the granularity-aware mask refiner to estimate precise instance mask, guided by the precision factor for precise spatial control over instance regions.
    
    \item We design the self-feedback audio agent as a separate-generate-remix-rework pipeline, curating accompanying audios to provide explicit temporal control for videos.
    
\end{itemize}

%% file: sec/2_related.tex
\section{Related work} \label{sec:related_work}

\subsection{Visual-only Video Generation and Editing} 
With the rapid development of video foundation models~\cite{wan,cogvideox,hunyuan}, video generation models demonstrate the great potential for animation creation~\cite{animatediff}, advertising~\cite{bara2022artificial}, cinematic visual effects~\cite{zhang2025generative, zhang2025stage}, and game development~\cite{genie}. 
This has motivated researchers to explore high-freedom editing techniques. 
Early editing methods~\cite{rave, vidtome, tune-a-video, fatazero} primarily adapt pre-trained text-to-image models~\cite{sd, flux}. However, as these backbones lack dynamic video priors, they inevitably struggle with temporal consistency. 
To address this, recent works~\cite{videopainter, gencompositor, vires, vace} propose dedicated video editing architectures. These methods achieve higher-quality and more consistent edits by incorporating specialized mechanisms (\eg, dual-branch encoders~\cite{videopainter, gencompositor} and auxiliary conditions~\cite{vires, vace}) for precise spatial control. 
However, these methods are fundamentally visual-only. They entirely neglect the accompanying audio cues, thereby breaking the original audio-visual synchronization that is crucial for an engaging video experience.

\subsection{Audio-guided Image Generation and Editing} 
Audio conveys rich semantic and affective cues that can serve as a powerful conditioning signal. Early audio-guided works focus on specific domains (\eg, music~\cite{sound2sight} and speech~\cite{speech2face}). With the emergence of diffusion models, these works begin to project audio into a shared embedding space to align audio semantics with visual content. Lee~\etal~\cite{lee2023generating} propose the audio attention and sentence attention to represent the rich audio characteristics. AudioToken~\cite{audiotoken} introduces an audio embedder to transform audio into text tokens used as conditioning for the pretrained text-to-image model~\cite{sd}. However, constrained by the static nature of images, these methods reduce audio to a holistic semantic feature. They inherently overlook the temporal cues and the frame-by-frame synchronization natural to the visual world.

\subsection{Audio-sync Video Generation and Editing} 
Modeling the natural synchronization between audio and video is a long-standing challenge. Early approaches achieve initial success for audio-sync video generation~\cite{tats, lee2022sound, Träumerai} based on GANs~\cite{vqgan, stylegan}. 
Recent generation models are built upon diffusion architecture~\cite{uniVerse-1, ovi}, which use multi-modal cross-attention mechanisms to naturally synthesize synchronized audios. 
In contrast, editing models extract fundamental audio signals (\eg, magnitude modulation~\cite{audioscenic} and distance guidance~\cite{Soundini}) to modify the global semantics of videos. Towards frame-by-frame temporal synchronization, recent works~\cite{aved, objavedit} invert the pretrained generation models and edit the latent code in a training-free manner. 
Despite this, these approaches still struggle to achieve full control (\eg, instance-level editing with free-form temporal controllability). Therefore, we propose AVI-Edit to bridge this gap by introducing the granularity-aware mask refiner and the self-feedback audio agent.

%% file: fig/pipeline.tex
\begin{figure*}[t]
  \centering
  \includegraphics[width=\linewidth]{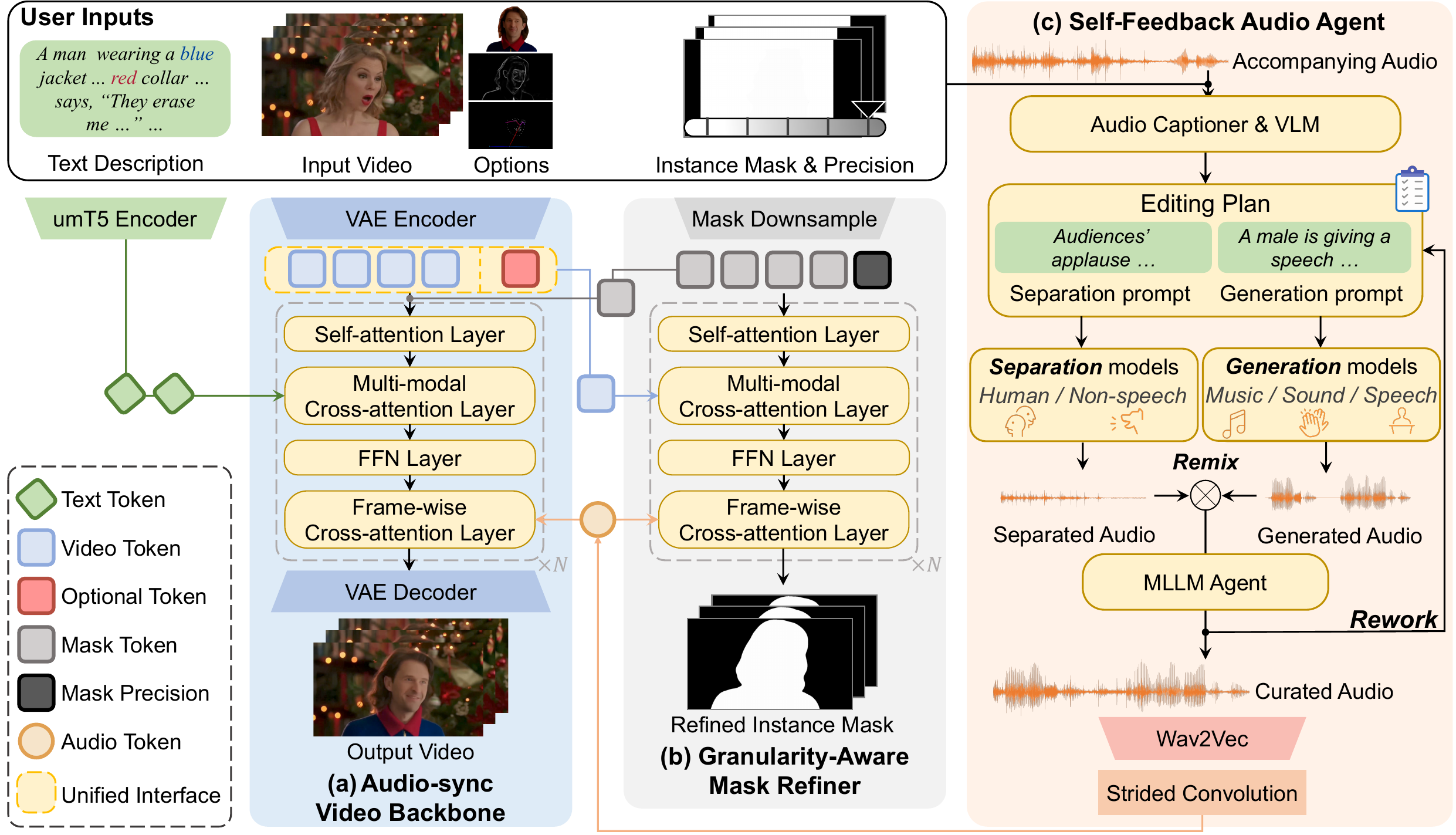}
  \vspace{-7mm}
  \caption{Illustration of our AVI-Edit framework.
  Given multi-modal user inputs, AVI-Edit separately encodes them into latent tokens, and generates audio-sync video instance editing results with the following components:  
  (a) The audio-sync video backbone (Sec.~\ref{sec:video_editing}) is built upon the pretrained video diffusion transformer, which utilizes a frame-wise cross-attention layer to understand audio cues and generates editing results based on the masks, text, and optional contexts.
  (b) The granularity-aware mask refiner (Sec.~\ref{sec:mask_refiner}) features a similar architecture and refines the coarse instance mask into a precise one, which utilizes visual semantics by replacing text tokens with video tokens, and is guided by the precision factor and audio cues.
  (c) The self-feedback audio agent (Sec.~\ref{sec:audio_agent}) is designed with a separate–generate–remix–rework pipeline that integrates off-the-shelf audio components to robustly handle diverse scenarios, which provides the resulting curated audio tokens as the temporal guidance for the video and mask components.
  }
  \label{fig:pipeline}
  \vspace{-4mm}
\end{figure*}

%% file: sec/3_dataset.tex
\section{Dataset} \label{sec:dataset}

While existing datasets~\cite{vggsound,audioset, lee2022sound} provide large-scale audio-visual correspondence, they typically lack the instance-level focus. 
To facilitate model training and evaluation, we construct \textbf{\sc AVISet}, a new dataset providing instance-centric correspondence and comprehensive annotations, tailored for the audio-sync video instance editing.

\noindent \textbf{Data sources.} 
We construct our initial data pool from two main sources: \textit{(i)} 5,529 hours of publicly available videos, including MovieBench~\cite{moviebench} (69h), Condensed Movies~\cite{cdm} (1,270h), Short-Films-20K~\cite{sf20k} (3,582h), and VGGSound~\cite{vggsound} (608h); and \textit{(ii)} 5,607 hours of open-access online videos (\eg, from YouTube). All collected videos retain their accompanying audios.

\noindent \textbf{Video filtering.}
Following prior work~\cite{midas, interacthuman,svd}, we first use PySceneDetect~\cite{pyscenedetect} to segment the original videos into single-shot clips.
We then use RAFT~\cite{raft} to compute optical flow to discard static or low-motion clips.
The aesthetic predictor~\cite{ase} is further adopted to measure the visual appeal and filter out clips with low visual quality.

\noindent \textbf{Audio filtering.}
We first use Audiobox-aesthetics~\cite{audiobox-aesthetics} to assess the quality of the accompanying audios, discarding low-quality clips.
Next, we adopt Qwen-Omni~\cite{qwen-omni} to classify all remaining clips into human speech or non-speech event clips for content-specific filtering.
For human speech clips, we apply TalkNet~\cite{talknet} to detect the active speaker's bounding box, and adopt Scribe~\cite{scribe} to identify speech segments of each speaker, allowing us to discard clips with multiple simultaneous speakers.
For non-speech event clips, we further use Qwen-Omni~\cite{qwen-omni} to filter out clips containing multiple sounding instances, and then predict the semantic category label (\eg, dog barking) for the single instance in the retained clips.

\noindent \textbf{Instance masking.}
Using the speaker's bounding box or the event's semantic category, we employ Grounded-SAM-2~\cite{grounded-sam2} to generate instance masks for each clip.
We further discard samples with small or empty mask regions, thereby filtering out cases where the sounding instance is not clearly visible or is off-screen (\eg, a voice-over).
After this stage, each video in the pool contains only one primary sounding instance, paired with its corresponding instance masks. All retained clips (for both human speech and non-speech event paths) are then pooled together, ensuring the final dataset is mixed for a general-purpose generation model.

\noindent \textbf{Text annotations.}
We adopt Qwen-VL~\cite{qwen-vl} to generate a scene-level text description for each retained clip.
For the testing set, we provide additional paired original–edited text instructions. Specifically, we prompt the LLM~\cite{qwen} with the clip's original text description and its predicted semantic category to generate a plausible editing instruction.
Finally, all generated instruction pairs are manually verified by volunteers to ensure their naturalness.

\noindent \textbf{Data statistics.}
The {\sc AVISet} includes 71k training, 1k validating, and 1k testing clips, totaling over 197 hours. Each clip is approximately 10 seconds, presented at 720P resolution (24 FPS).
We annotate each clip with an instance mask and a scene-level text description, while the testing set additionally includes paired original–edited text instructions.

%% file: sec/4_method.tex
\section{Methodology} \label{sec:methodology}
AVI-Edit consists of three main components. 
We first introduce the audio-sync video backbone (\cref{sec:video_editing}) with its architecture and formulation.
We then detail the granularity-aware mask refiner (\cref{sec:mask_refiner}), which provides precise spatial control over the edited instance. 
Finally, we describe the self-feedback audio agent (\cref{sec:audio_agent}), which generates fine-grained temporal guidance for the entire editing process.

\subsection{Audio-Sync Video Backbone} \label{sec:video_editing}
As shown in~\cref{fig:pipeline} (a), the audio-sync video backbone generates audio-sync video editing results based on a curated audio $a$, an instance mask $m$, and a text description $y$.

\noindent \textbf{Video compression.} 
To enable editing in the compact latent space, we encode the original clip $x$ with a pretrained VAE encoder $\mathcal{E}$ into a clean latent code as $z=\mathcal{E}(x)$.
A corresponding VAE decoder $\mathcal{D}$ is used to reconstruct the video clip from the latent code as $x = \mathcal{D}(z)$.

\noindent \textbf{Flow matching formulation.}
We train our video diffusion transformer $v_{\theta}$ using the flow matching objective~\cite{flow_matching}, and define the probability path $\hat{z}_t$ at timestep $t$ as a linear combination of the latent code $z$ and Gaussian noise $\epsilon \sim \mathcal{N}(0, \mathbf{I})$:
\begin{equation} \label{eq:flow_matching_add_noise}
\hat{z}_t = tz + (1 - t)\epsilon,
\end{equation}
where $t \in [0, 1]$ is the timestep, with $\hat{z}_0 = \epsilon$ and $\hat{z}_1 = z$.

\noindent \textbf{Background preservation.}
To avoid producing undesirable alterations and inconsistencies in background regions, we adopt the instance mask $m$ to indicate the edited regions, and downsample it to compose the interpolated $\hat{z}_t$ with the clean input video tokens $z$:
\begin{equation}
\label{eq:latentmask}
z_t = \hat{z}_t \odot \hat{m} + z \odot (1-\hat{m}),
\end{equation} 
where $\hat{m}$ is the downsampled instance binary mask.

\noindent \textbf{Audio-sync diffusion transformer.}
To leverage priors of the pretrained video generation model, we build the video backbone upon the Wan2.2 architecture~\cite{wan}.
We first concatenate the downsampled instance mask $\hat{m}$ with the noisy latent code $z_t$ as video tokens. After that, these tokens are sequentially processed by a self-attention layer to interact with long-range context, a multi-modal cross-attention layer to understand the text description, and a feed-forward network to extract features for $N$ Diffusion Transformer (DiT) blocks.
To enable audio-sync video editing, we additionally introduce a frame-wise cross-attention to each DiT block, which receives curated audio tokens $a$ and guides the temporal alignment with the video latent codes.

\noindent \textbf{Optional control contexts.}
To enable video editing results precisely satisfying the user's intention, we design a unified interface to incorporate optional control contexts $c$ (\eg, scribble, pose, and reference image). 
These conditions are encoded to context tokens using the pretrained VAE encoder $\mathcal{E}$, and then injected through element-wise addition for scribble and pose, and concatenation for the reference image. This provides fine-grained controllability.

\noindent \textbf{Training and inference process.}
During training, we finetune the video diffusion transformer $v_{\theta}$ to estimate the velocity field $v_t = z - \epsilon$, by minimizing the standard flow matching objective:
\begin{equation} \label{eq-fm}
\mathcal{L}_\mathrm{fm} =  \mathbb{E}_{\epsilon,t,z,\hat{m},y,a,c} \big[ \left\| v_{\theta}(z_t, t, \hat{m}, y, a, c) - v_t \right\|^2 \big].
\end{equation}
During inference, we transform the Gaussian noise $\epsilon$ into the edited clean latent code $z_1$ by solving:
\begin{equation} \label{eq-ode}
    \textrm{d}z_t/\textrm{d}t = v_{\theta}(z_t, t, \hat{m}, y, a, c),
\end{equation}
from $t = 0$ to $t = 1$ using a numerical solver.

\subsection{Granularity-Aware Mask Refiner} \label{sec:mask_refiner}
The instance mask allows users to specify the target instance to be edited.  However, user-provided instance masks are typically inaccurate (\eg, bounding box). As presented in \cref{fig:pipeline}~(b), we propose the Granularity-Aware Mask Refiner (GAMR) to refine these coarse instance mask, thereby achieving precise instance-level video editing.

\noindent \textbf{Mask granularity.}
To quantitatively describe the granularity of instance mask, we first introduce a precision factor $p \in [0, P]$, which measures the degree of alignment between the target instance region and the provided mask. Conceptually, $p = P$ corresponds to maximal degradation (\eg, a bounding box), while $p = 0$ denotes the precise instance contour $m_\mathrm{gt}$.

\noindent \textbf{Diffusion-based refinement.}
We implement the GAMR with a similar architecture of video diffusion transformer, designed to predict the accurate instance mask.
To understand the granularity of the instance mask, the precision factor $p$ is linearly encoded and injected into GAMR through the AdaLN and Gate in each diffusion transformer block.
The curated audio tokens are also incorporated via frame-wise cross-attention, enabling the GAMR to infer instance masks aligned with event timing. 
In contrast, the text tokens in the multi-modal cross-attention layers are replaced with video tokens, which enables the GAMR to further reason about instance masks based on the visual semantics.

\noindent \textbf{Training strategy.}
To simulate the inaccuracy of user-provided masks, we sample input videos $x$ and their corresponding precise instance masks $m_\mathrm{gt}$ from our {\sc AVISet}, then degrade each mask to an arbitrary granularity level $p$ to construct training pairs. 
Specifically, the degradation is implemented using a set of predefined Gaussian blur kernels:
\begin{equation} \label{eq:mask_degradation}
m_p = \mathrm{GaussianBlur}(m_\mathrm{gt}, k_p, \sigma_p),
\end{equation}
where the kernel size $k_p$ and standard deviation $\sigma_p$ are determined by the precision factor $p$. 
We introduce a mask refinement objective~\cite{focal_loss} that adopts focus loss to measure the discrepancy between the downsampled generated mask $\hat{m}$ and ground-truth mask $\hat{m}_\mathrm{gt}$ to train the GAMR:
\begin{equation} \label{eq:loss_mask}
\begin{aligned}
    \mathcal{L}_{\mathrm{mask}} = & -\,\alpha\, \hat{m}_\mathrm{gt}(1-\hat m)^{\gamma}\log(\hat m) \\
                   & -\,(1-\alpha)(1-\hat{m}_\mathrm{gt})\,\hat m^{\gamma}\log(1-\hat m),
\end{aligned}
\end{equation}
where $\alpha = 0.25$ and $\gamma = 2.0$ are hyperparameters.

\noindent \textbf{Inference process.}
The GAMR estimates the refined mask $\hat{m}$ at each step of the ODE solving process. 
Specifically, this iterative refinement is precision-aware for each inference step.
At the first step ($k=0$), we provide the downsampled user-provided mask $\hat{m}_p$ as $\hat{m}^0_p$ and its precision factor $p$ (\eg, $p=P$ for a bounding box).
For all subsequent steps ($k > 0$), we iteratively feed the estimated mask $\hat{m}^{k-1}_p$ from the previous step as the new input mask $\hat{m}^k_p$. 
The precision factor $p$ for these steps is adjusted according to a predefined degradation schedule.
At each step $k$, the concurrently refined mask $\hat{m}^{k}_p$ is then used as the definitive mask $\hat{m}$ for the video generation backbone (\cref{eq-ode}).

\input{fig/comparison}

\subsection{Self-Feedback Audio Agent} \label{sec:audio_agent}
To provide the temporal guidance for audio-sync video editing, we design a dedicated audio agent to curate the original audio $a_\mathrm{orig}$.
As illustrated in \cref{fig:pipeline}~(c), the audio agent is guided by the text description $y$ and the associated visual context. 
It is designed with a separate–generate–remix–rework pipeline to integrate a collection of off-the-shelf audio processing components, thereby enabling robust handling of diverse scenarios (\eg, human speech and non-speech events).

\noindent \textbf{Audio separation and generation.}
To obtain a comprehensive understanding of the original audio $a_\mathrm{orig}$, we employ a captioner~\cite{qwen-omni} to translate the audio into a detailed text description and summarize its semantic content $c_\mathrm{sem}$ (\eg, sound events and environment atmosphere). Next, we leverage a VLM~\cite{qwen-vl} to reason about the detailed editing plan:
\begin{equation} \label{eq:audio_agent_neg_pos}
    (c^\mathrm{sep}, c^\mathrm{gen}) = \mathrm{VLM}([x, m_p, c_\mathrm{sem}, y]),
\end{equation} 
where $c^\mathrm{sep}$ and $c^\mathrm{gen}$ are text descriptions for the audio component to be remained from the separation (\eg, applause) and to be generated (\eg, male speech), respectively.
Guided by this editing plan, the agent parses these descriptions to select the most suitable separation ($T^\mathrm{sep}_s$) and generation ($T^\mathrm{gen}_g$) models among a collection of off-the-shelf models, and applies them to obtain the remaining ($a^\mathrm{sep}$) and generated ($a^\mathrm{gen}$) components:
\begin{equation} \label{eq:audio_agent_rem_gen}
    a^\mathrm{sep} = T^\mathrm{sep}_s(a_\mathrm{orig}, c^\mathrm{sep}), \quad  a^\mathrm{gen} = T^\mathrm{gen}_g(c^\mathrm{gen}).
\end{equation}

\noindent \textbf{Model collection.}
As human speech is best handled by specialized models, we define separation component as a collection of domain-specific models (\ie, human speech~\cite{Spleeter} and non-speech events~\cite{omnisep}) as $T^\mathrm{sep} \!=\! \{T^\mathrm{sep}_s\}_{s \in \mathcal{S}}$, where the domain set is $\mathcal{S} \!=\! \{\text{speech, non-speech}\}$. 
As audio can also be decomposed into distinct tracks~\cite{cdx}, we define our generation components as $T^\mathrm{gen} \!=\! \{T^\mathrm{gen}_g\}_{g \in \mathcal{G}}$,  where the track set is $\mathcal{G} \!=\! \{\text{speech, music, sound}\}$. These are implemented with text-to-speech, text-to-music, and text-to-sound models from ElevenLabs~\cite{elevenlabs}.

\noindent \textbf{Remix and rework judgment.} 
After obtaining the separated component $a^\mathrm{sep}$ and generated component $a^\mathrm{gen}$, we remix them to produce the curated audio $a$~\cite{pydub}. 
To prevent the degraded editing quality, we employ a multimodal large language model (MLLM)~\cite{qwen-omni} to evaluate the overall perceptual quality $q$ of the remixed curated audio (\eg, whether it sounds natural and realistic). 
Only remixed audios with quality score exceeding a predefined threshold $\tau$ are accepted. 
When the judging MLLM rejects a remixed result, it generates an improvement instruction $(\hat{c}^\mathrm{sep}, \hat{c}^\mathrm{gen})$, providing feedback to both the separation model $T^\mathrm{sep}$ (\eg, female speech remains audible) and the generation model $T^\mathrm{gen}$ (\eg, synthesized male speech volume is insufficient), enabling an iterative refinement loop.  The rework loop continues until the remixed audio’s quality score surpasses $\tau$ or a maximum number of iterations is reached.

%% file: fig/comparison.tex
\begin{figure*}[t]
  \centering
  \hfill
  \includegraphics[width=\linewidth]{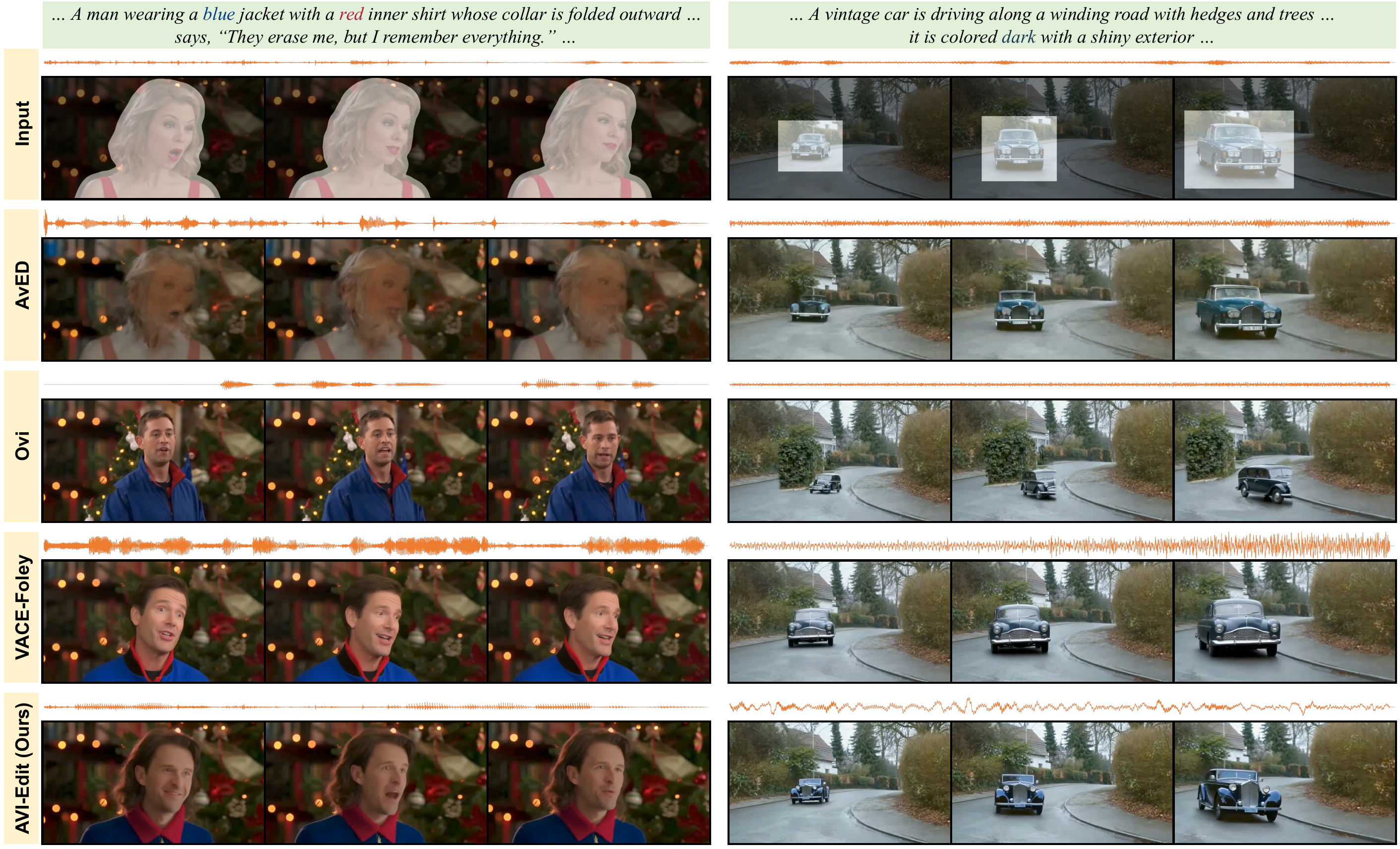}
  \vspace{-7mm}
  \caption{Qualitative comparison results with state-of-the-art methods for audio-sync video instance editing.}
  \vspace{-5mm}
  \label{fig:comparison}
\end{figure*}

%% file: sec/5_experiment.tex
\section{Experiment}

\noindent \textbf{Training details.}
We initialize our video backbone and GAMR with pre-trained weights from Wan2.2-5B~\cite{wan}, freezing the spatial-temporal VAE and fine-tuning the diffusion Transformer to minimize the joint flow matching objective (\cref{eq-fm}) and mask refiner objective (\cref{eq:loss_mask}):
\begin{equation} \label{eq:loss_final}
\mathcal{L} = \mathcal{L}_\mathrm{fm} + \lambda \mathcal{L}_\mathrm{mask},
\end{equation}
where the hyperparameter $\lambda=1.0$ in our experiments.
We train our AVI-Edit framework for 160k steps on 8 NVIDIA A800 GPUs, using a spatial resolution of 720p. We use the Adam optimizer~\cite{adam} with a learning rate of $2 \times 10^{-5}$.

\noindent \textbf{Evaluation Datasets.} 
In addition to our proposed {\sc AVISet} for the general audio-sync video editing task, we further evaluate our AVI-Edit framework on the publicly available AvED-Bench~\cite{aved} to demonstrate its superior performance and generalization capability. 
To enable effective instance-level evaluation on AvED-Bench, we annotate instance masks according to the procedure described in \cref{sec:dataset}. 
For evaluation, we randomly sample 100 video clips from each dataset's test split. 

\input{tab/comparison}

\subsection{Comparison with State-of-the-art Methods} 
As audio-sync video instance editing is an emerging research task, the comparison methods are still under development. We compare our framework with two recent state-of-the-art methods, AvED~\cite{aved} and Ovi~\cite{ovi}. 
We re-implement AvED~\cite{aved} strictly following its original settings, due to its code being unavailable. 
As Ovi~\cite{ovi} is a generation model, we adapt it for editing using a common zero-shot inpainting strategy, where only the mask region is replaced with Gaussian noise. 
Furthermore, we implement a sequential baseline (VACE-Foley), where VACE~\cite{vace} edits the video first, followed by Hunyuan-Foley~\cite{hunyuan-foley} to generate the accompanying audio.  All baselines are fine-tuned on the {\sc AVISet}.

\input{tab/comparison_user_study}

\input{tab/audio_eval}

\noindent \textbf{Quantitative comparisons.} 
We quantitatively evaluate performance across three main aspects: 
\textit{(i)} Visual quality is examined using Fréchet Video Distance (FVD)~\cite{fvd} and Inception Score (IS)~\cite{is}. 
\textit{(ii)} Frame consistency (FC) is measured by calculating similarity between consecutive frames using CLIP~\cite{clip}. 
\textit{(iii)} Alignment is assessed via text-video (TC) with CLIP-XL~\cite{videoclipxl}, audio-video (AC) with ImageBind~\cite{imagebind}, and lip motion synchronization (\ie, Sync-C and Sync-D)~\cite{sync}. 
Notably, since AvED-Bench~\cite{aved} does not contain human talking videos, the lip motion synchronization metrics are not applicable for it. 
Since the baseline methods are not designed to handle coarse masks, we evaluate all methods using ground-truth instance masks and a text description to ensure a fair comparison.
As presented in~\cref{tab:comparison}, our framework demonstrates significant advantages on both datasets, outperforming state-of-the-art methods on most quantitative metrics. 
Metrics details are provided in the supplementary material.

\noindent \textbf{Qualitative comparisons.} 
We present qualitative comparisons with the aforementioned baselines~\cite{aved, ovi, vace, hunyuan-foley} to simulate a more realistic use case. For this evaluation, all methods are provided with the coarse mask and text description.
As shown in \cref{fig:comparison}, the left and right columns evaluate the case of human speech and non-speech events, respectively. 
As a result, AvED~\cite{aved} struggles to maintain temporal consistency, resulting in noticeable flickering artifacts (\cref{fig:comparison} left and right). 
Ovi~\cite{ovi} suffers from visual inconsistencies (\eg, \cref{fig:comparison} right), while VACE-Foley~\cite{vace,hunyuan-foley} fails in synthesizing the target speech (\cref{fig:comparison} left).
In contrast, our AVI-Edit framework produces visually appealing video instance editing results with realistic accompanying audios.

\input{fig/ablation}

\input{tab/ablation}

\input{fig/application}

\noindent \textbf{User preference study.}
We additionally conduct a user study to evaluate human preference. Given ground-truth instance masks and text instructions, AVI-Edit and relevant methods produce their edit results. Participants are asked to evaluate these results from three aspects: 
\textit{(i)} \textbf{Audio-Visual Synchronization (AVS):} Participants are asked to select the video clip that is the best synchronized with the accompanying audio. 
\textit{(ii)} \textbf{Text Alignment (TA):} Participants are asked to select the video clip that best matches the input text instruction. 
\textit{(iii)} \textbf{Overall Preference (OP):} Participants are asked to select the video clip they prefer overall. 
We randomly selected 10 samples from each dataset, and recruited 25 volunteers to provide independent evaluations across all three aspects. As shown in \cref{tab:compare-user-study}, our model achieves the highest preference scores in all aspects.

\noindent \textbf{Audio quality study.}
To assess the quality of audios produced by our self-feedback audio agent, we conduct an additional user study. We run our agent on 50 random clips from {\sc AVISet}, and recruit 25 volunteers to evaluate these results on the following three aspects:
\textit{(i)} \textbf{Audio Fidelity (AF):} to evaluate the realism and clarity of the resulting audio; \textit{(ii)} \textbf{Remaining Preservation (RP):} to assess whether the non-target audio components are naturally preserved; and 
\textit{(iii)} \textbf{Text-Audio Consistency (TAC):} to measure the semantic alignment between the result audio and the user-provided text description.
Volunteers rate each sample using a 4-point scale: ``Failed", ``Borderline", ``Acceptable", or ``Perfect". As shown in \cref{tab:audio-evaluation}, our agent performs strongly, with over 91\% (AF), 85\% (RP), and 88\% (TAC) of ratings being ``Acceptable" or ``Perfect".

\subsection{Ablation Study} 
We conduct ablation studies to evaluate the impact of key components of AVI-Edit. 
During ablation, the instance mask is randomly degraded into a coarse one to evaluate the robustness of the mask refinement.
The evaluation scores and editing results of the ablation study are presented in \cref{tab:ablation} and \cref{fig:ablation}, respectively.

\noindent \textbf{W/o Precision Factor (PF).}
This variant discards the precision factor that indicates the mask granularity, causing the granularity-aware mask refiner to lose its granularity guidance. This variant results in an imprecise estimation of the woman's head regions within the mask (\cref{fig:ablation}, second row).

\noindent \textbf{W/o Mask Refiner (MR).}
This variant removes the granularity-aware mask refiner, forcing the model to rely solely on the initial coarse mask. As a result, this variant struggles to faithfully preserve background regions, altering the wall behind the head (\cref{fig:ablation}, third row).

\noindent \textbf{W/o Audio Agent (AA).} 
We replace our self-feedback audio agent with a general audio editing model~\cite{audioeditcode}. This leads to noisier audio guidance and degraded audio-visual synchronization (\cref{tab:ablation}, Audio-C and Sync-C scores).

\subsection{Application} 
As illustrated in \cref{sec:video_editing}, our video backbone is equipped with a unified interface that supports diverse, \textit{optional} control contexts beyond the primary text and audio guidance.
We demonstrate this additional capability in \cref{fig:application},  where users can draw the scene with the scribble, manipulate the instance's gesture with the pose, and customize the instance's appearance with the reference image. 
This controllability allows users to satisfy highly specific intentions.
Due to space limitations, we provide further demonstrations in the supplementary material (\eg, instance insertion and removal, long video editing, and audio-sync generation).

%% file: tab/comparison.tex
\begin{table*}[t]
\caption{Quantitative experiment results of comparison with state-of-the-art methods. $\uparrow$ ($\downarrow$) means higher (lower) is better. Throughout the paper, best performances are highlighted in \textbf{bold}.} 
\vspace{-6mm}
\begin{center}
{
    \setlength\tabcolsep{4pt}
    \centering
    \begin{adjustbox}{width={\textwidth},totalheight={\textheight},keepaspectratio}
    \begin{tabular}{l | c @{\hspace{15pt}} c c c c c c| c @{\hspace{15pt}} c c c c}  \toprule
    \multirow{2}{*}{Method} & \multicolumn{7}{c|}{\sc AVISet} & \multicolumn{5}{c}{AvED-Bench} \\
    & FVD $\downarrow$ & IS $\uparrow$ & FC (\%) $\uparrow$  & TC (\%) $\uparrow$  & AC (\%) $\uparrow$ & Sync-C $\uparrow$ & Sync-D $\downarrow$ & FVD $\downarrow$ & IS $\uparrow$ & FC (\%) $\uparrow$  & TC (\%) $\uparrow$  & AC (\%) $\uparrow$ \\ \midrule
    
    AvED & 364.69 & 1.104 & 95.03 & 23.69 & 23.31 & 1.69 & 11.80 & 413.82 & 1.118 & 94.89 & 24.59 & 20.45 \\

    Ovi & 407.08 & 1.122 & 96.47 & 25.83 & 26.68 & 4.00 & \textbf{9.12} & 504.14 & 1.122 & 95.65 & 25.21 & 21.49 \\

    VACE-Foley & 391.64 & 1.115 & 96.60 & 25.92 & 26.45 & 1.72 & 10.37 & 405.76 & 1.108 & 95.74 & 25.17 & 21.46 \\
    
    AVI-Edit (Ours) & \textbf{312.89} & \textbf{1.127} & \textbf{96.65} & \textbf{26.16} & \textbf{26.93} & \textbf{4.12} & 9.19 & \textbf{349.31} & \textbf{1.125} & \textbf{95.82} & \textbf{25.30} & \textbf{21.64} \\
    \bottomrule
    
    \end{tabular}\label{tab:comparison}
    \end{adjustbox}
}
\end{center}
\vspace{-7mm}
\end{table*}

%% file: tab/comparison_user_study.tex
\begin{table}[t] 
\caption{Percentage (\%) of user preference study results.}
\vspace{-2mm}
\label{tab:compare-user-study}
\centering 
\footnotesize
\setlength\tabcolsep{4pt} 
\begin{tabular}{l | ccc | ccc} 
\toprule 
\multirow{2}{*}{Method} & \multicolumn{3}{c|}{\sc AVISet} & \multicolumn{3}{c}{AvED-Bench} \\
& AVS & TA & OP & AVS & TA & OP \\ 
\midrule
AvED & 2.40 & 3.20 & 1.60 & 3.60 & 4.80 & 4.00 \\
Ovi & 36.00 & 36.80 & 38.40 & 31.60 & 31.20 & 32.00 \\
VACE-Foley & 12.40 & 17.20 & 14.80 & 19.20 & 21.60 & 22.80 \\
AVI-Edit (Ours) & \textbf{49.20} & \textbf{42.80} & \textbf{45.20} & \textbf{45.60} & \textbf{42.40} & \textbf{41.20} \\
\bottomrule
\end{tabular}
\vspace{-1mm}
\end{table}

%% file: tab/audio_eval.tex
\begin{table}[t]
\caption{Percentage (\%) of audio quality study results.} 
\vspace{-6mm}
\label{tab:audio-evaluation} 

\begin{center}
{
    \setlength\tabcolsep{14pt}
    \centering
    \begin{adjustbox}{width={0.40\textwidth},totalheight={\textheight},keepaspectratio}
    \begin{tabular}{l | c c c}  \toprule
    Rating & AF & RP & TAC \\ \midrule
     Failed & 2.24 & 5.36 & 3.76 \\
    Borderline & 6.32 & 9.12 & 7.60 \\
    Acceptable & 8.48 & 19.68 & 14.96 \\
    Perfect & \textbf{82.96} & \textbf{65.84} & \textbf{73.68} \\
    \bottomrule
    
    \end{tabular}
    \end{adjustbox}
}
\end{center}
\vspace{-8mm}
\end{table}

%% file: fig/ablation.tex
\begin{figure}[t]
  \centering
  \hfill
  \includegraphics[width=\linewidth]{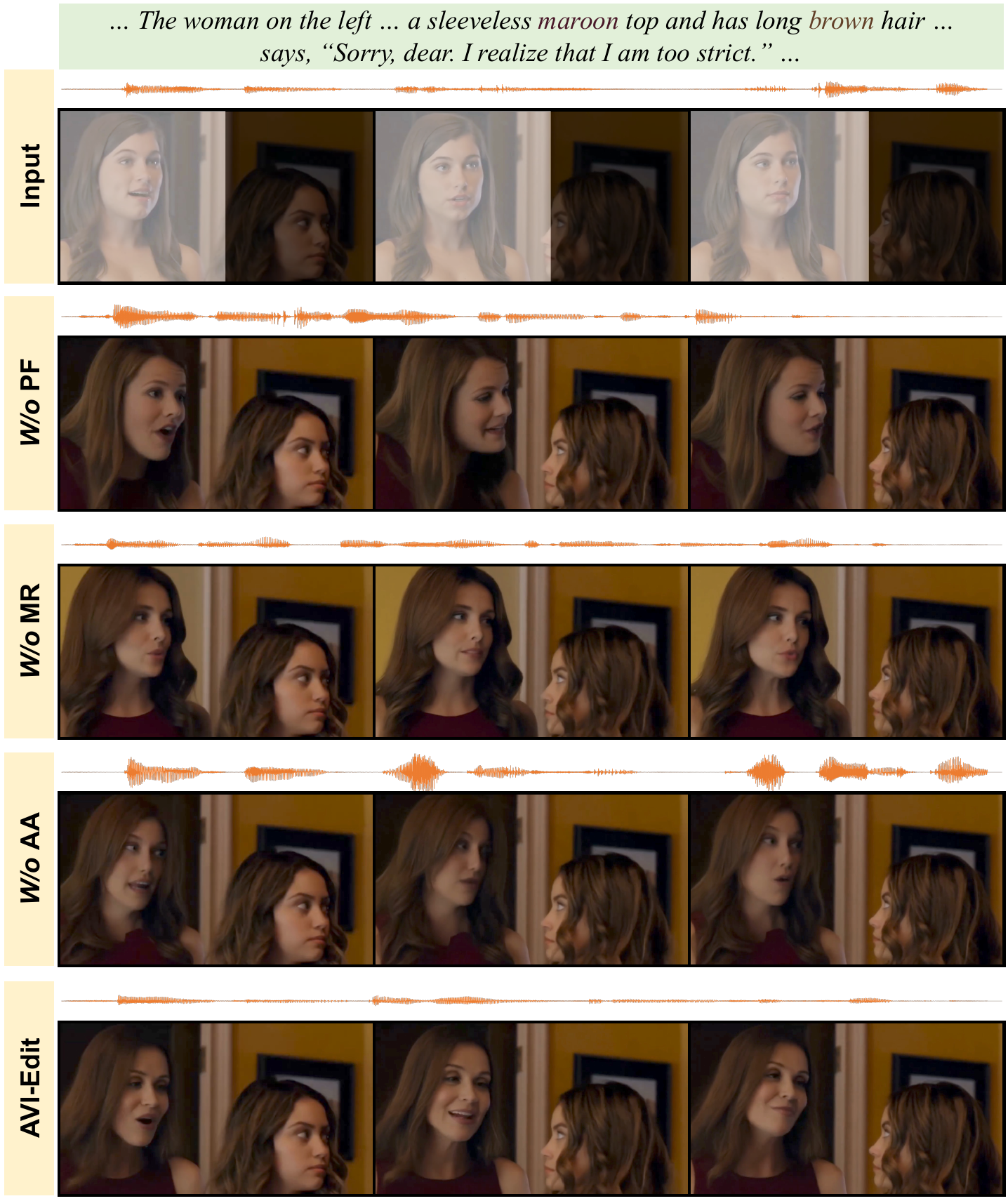}
  \vspace{-7mm}
  \caption{Visual results of ablation study with baseline variants.}
  \label{fig:ablation}
  \vspace{-5mm}
\end{figure}

%% file: tab/ablation.tex
\begin{table*}[t]
\caption{Quantitative experiment results of ablation studies. $\uparrow$ ($\downarrow$) means higher (lower) is better.} 
\vspace{-6mm}
\label{tab:ablation}
\begin{center}
{
    \setlength\tabcolsep{4pt}
    \centering
    \begin{adjustbox}{width={\textwidth},totalheight={\textheight},keepaspectratio}
    \begin{tabular}{l | c @{\hspace{15pt}} c c c c c c| c @{\hspace{15pt}} c c c c}  \toprule
    \multirow{2}{*}{Method} & \multicolumn{7}{c|}{\sc AVISet} & \multicolumn{5}{c}{AvED-Bench} \\
    & FVD $\downarrow$ & IS $\uparrow$ & FC (\%) $\uparrow$  & TC (\%) $\uparrow$  & AC (\%) $\uparrow$ & Sync-C $\uparrow$ & Sync-D $\downarrow$ & FVD $\downarrow$ & IS $\uparrow$ & FC (\%) $\uparrow$  & TC (\%) $\uparrow$  & AC (\%) $\uparrow$ \\ \midrule
    
    \textit{w/o} PF & \text{354.43} & \text{1.119} & \text{96.49} & \text{26.07} & \text{26.50} & \text{4.12} & \text{9.43} & \text{490.92} & \text{1.118} & \text{95.47} & \text{25.06} & \text{21.51} 
    \\
        
    \textit{w/o} MR & \text{372.44} & \text{1.107} & \text{96.32} & \text{25.68} &\text{26.38} & \text{4.07} & \text{9.36} & \text{539.83} & \text{1.103} & \text{95.29} & \text{24.96} & \text{21.45}
    \\

    \textit{w/o} AA & \text{342.75} & \text{1.114} & \text{96.54} & \text{25.84} &\text{25.97} & \text{3.83} & \text{9.61} & \text{445.56} & \text{1.105} & \text{95.36} & \text{25.13} & \text{21.22} 
    \\
    
    AVI-Edit   & \textbf{335.32} & \textbf{1.121} & \textbf{96.63} & \textbf{26.13} & \textbf{26.77} & \textbf{4.18} & \textbf{9.27} & \textbf{402.74} & \textbf{1.122} & \textbf{95.58} & \textbf{25.17} & \textbf{21.63}
    \\ 
    \bottomrule
    
    \end{tabular}
    \end{adjustbox}
}
\end{center}
\vspace{-4mm}
\end{table*}

%% file: fig/application.tex
\begin{figure*}[t]
  \centering
  \hfill
  \includegraphics[width=\linewidth]{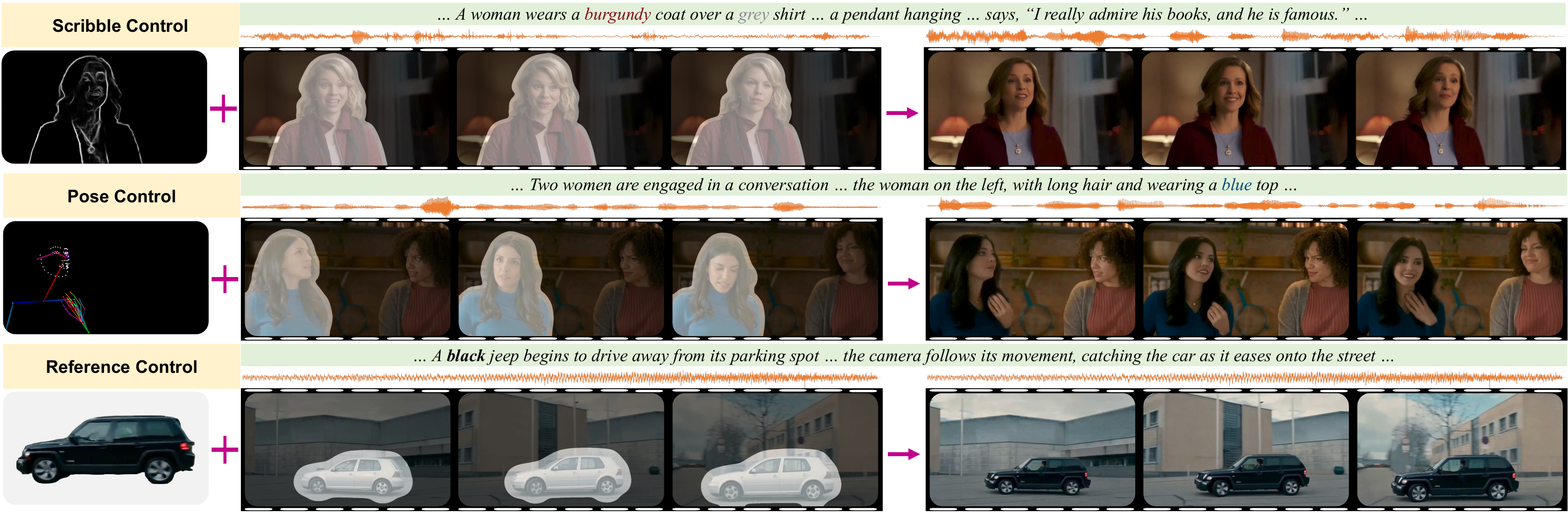}
  \vspace{-5mm}
  \caption{Versatile applications of AVI-Edit demonstrating its diverse controllability.}
  \vspace{-6mm}
  \label{fig:application}
\end{figure*}

%% file: sec/6_conclusion.tex
\section{Conclusion} 
In this paper, we present AVI-Edit, a framework designed for audio-sync video instance editing. 
We build the video generation backbone upon the pretrained text-to-video model~\cite{wan}, and equip it with frame-wise attention to receive audio guidance and a condition interface to edit the video content. 
We then introduce the precision factor to indicate the mask granularity, and adopt a similar architecture to build the granularity-aware mask refiner, which effectively refines the potentially coarse instance mask for accurate instance-level editing. We further propose the self-feedback audio agent, designed with a separate–generate–remix–rework pipeline to robustly handle scenarios of human speech and non-speech events. Extensive experiments demonstrate that AVI-Edit achieves state-of-the-art performance with diverse application scenarios.

\noindent \textbf{Limitation.} 
Since the target mask indicates an instance to be edited,  editing videos with multiple target instances requires running AVI-Edit sequentially, processing each instance one-by-one. 
We leave the exploration of simultaneous multi-instance editing in the future work.

\section*{Acknowledgements}
This work is supported by National Natural Science Foundation of China (Grant No. 62136001) and Beijing Major Science and Technology Project (Grant No. Z251100008125009). We thank all the insightful reviewers for the helpful suggestions, and the colleagues at Beijing Academy of Artificial Intelligence for their support.

%% file: sec/X_suppl.tex
\section{Appendix}

\begin{figure*}[t]
  \centering
  \hfill
  \includegraphics[width=\linewidth]{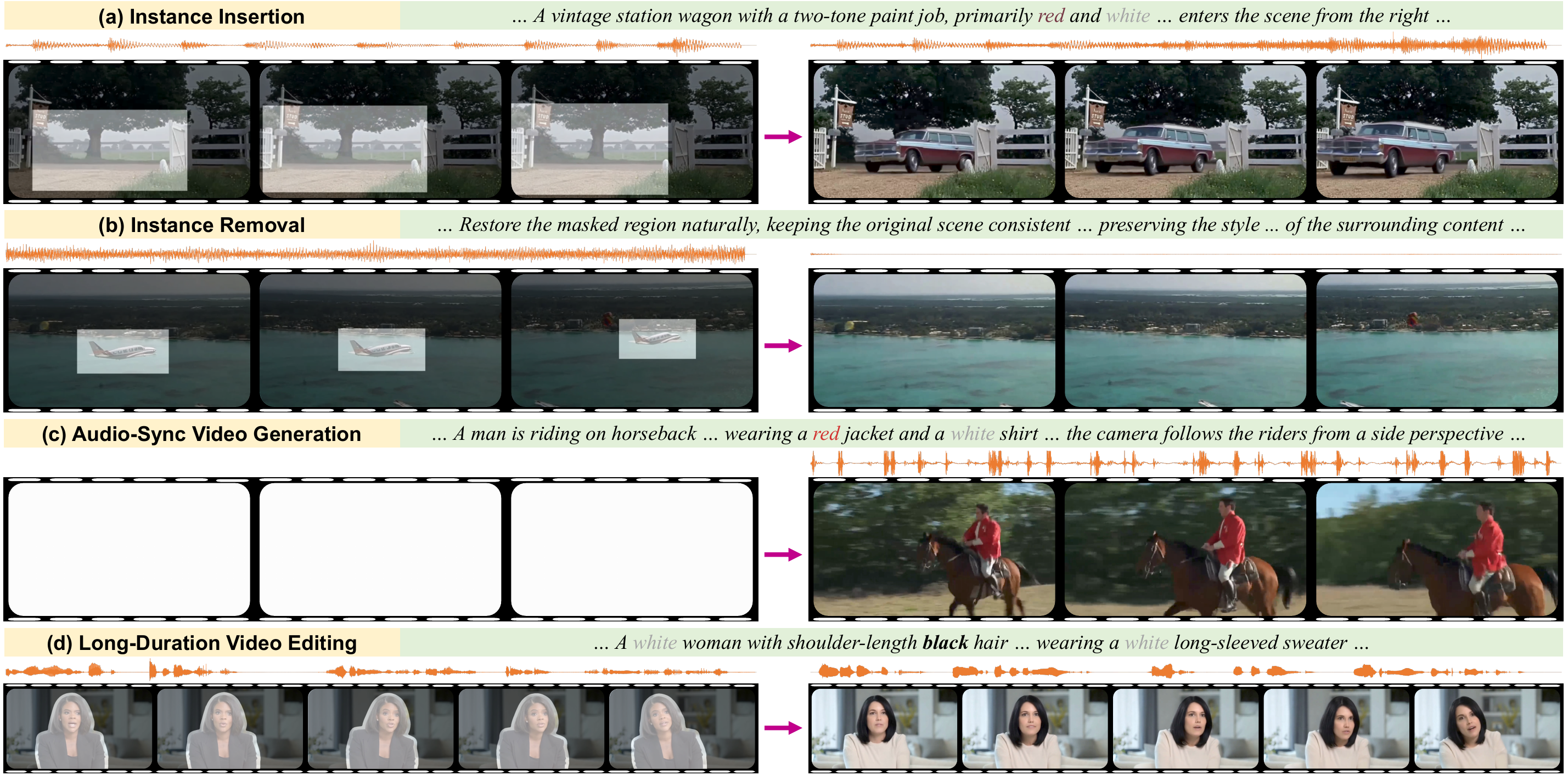}
  \caption{Additional application scenarios of AVI-Edit.}
  \label{fig:supp_application}
  \vspace{-3mm}

\end{figure*}

\let\thefootnote\relax
\footnotetext{
    \begin{minipage}[t]{\textwidth}
        $^\dag$ Equal contribution. \\
        $^\ddag$ Corresponding author.
    \end{minipage}
}
\subsection{Additional Application Scenarios} \label{sec:additional_application}
As illustrated in Sec.~\textcolor{cvprblue}{5.3}, we additionally present application scenarios in~\cref{fig:supp_application} to demonstrate the versatility of our framework.

\noindent \textbf{Instance insertion.} 
By specifying target regions within a user-provided video, AVI-Edit can generate the corresponding instance and its accompanying audio according to the text description, seamlessly composited with the background regions (\cref{fig:supp_application} first row, inserting a car with its accompanying audio into an empty field).

\noindent \textbf{Instance removal.} 
When an unwanted instance is specified, AVI-Edit removes the target instance and its accompanying audio from all frames based on the text description, resulting in a clean background (\cref{fig:supp_application} second row, removing a plane flying over the sea and its accompanying audio).

\noindent \textbf{Audio-sync video generation.} 
In the extreme case where the instance mask is expanded to cover the full video frame, AVI-Edit discards all video content and accompanying audios, becoming an audio-sync generation model according to the text description (\cref{fig:supp_application} third row, generating an audio-sync video from the provided text description).

\noindent \textbf{Long-duration video editing.}
To enable AVI-Edit to edit long-duration videos, we incorporate a conditioning strategy during training to unmask $k$ randomly selected frames. During inference, we generate the initial segment from noise. Subsequent segments are then generated using the last $k$ frames of the preceding clip as visual guidance, enabling the iterative editing of arbitrary-length videos (\cref{fig:supp_application} last row, consistently editing an instance spanning 241 frames).

\subsection{Precision Factor Details}
As illustrated in Sec.~\textcolor{cvprblue}{4.2}, as user-provided instance masks are typically inaccurate (\eg, bounding box), we introduce a precision factor $p$ to indicate the mask granularity.

In real applications, users typically provide the maximal precision factor $p=P$ for bounding boxes, or manually specify an intermediate $p$ value to provide precise control for the target instance region based on the provided mask.
To simulate the inaccuracy of user-provided masks, we randomly degrade each precise instance mask to an arbitrary granularity level $p$ during training (Eq.~(\textcolor{cvprblue}{5})).

\begin{figure*}[t]
  \centering
  \hfill
  \includegraphics[width=\linewidth]{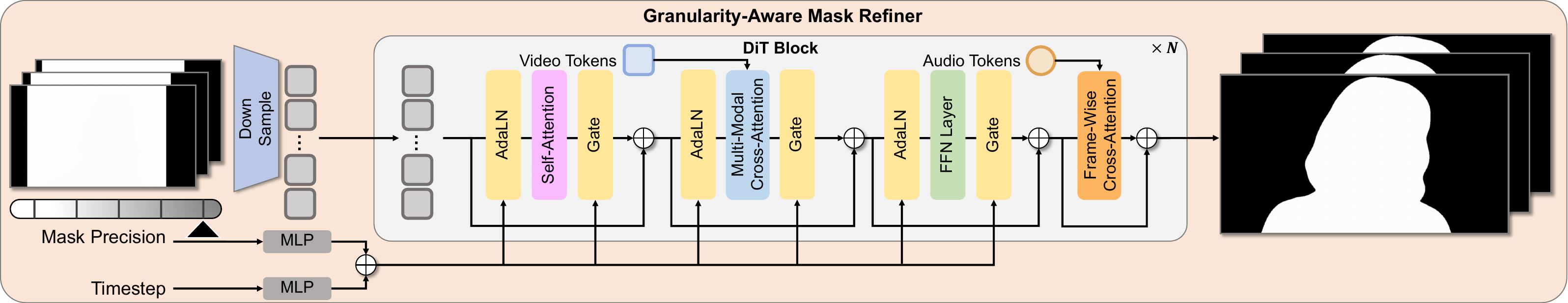}
  \caption{The detailed architecture of the granularity-aware mask refiner.}
  \label{fig:GAMR}
  \vspace{-3mm}
\end{figure*}

We further provide the architecture details of the Granularity-Aware Mask Refiner (GAMR) in \cref{fig:GAMR}, which demonstrates the approach of precision factor injection.
Specifically, the precision factor $p$ is linearly encoded and added to the timestep embedding to produce the modulation parameters:
\begin{align}
    f^\mathrm{p} & = \text{Linear}(p) + \text{Linear}(t), \\
    (\gamma, \beta, \alpha) & = \text{Linear}(f^{p}),
\end{align}
where $t$ is the timestep embedding and $(\gamma, \beta, \alpha)$ are estimated parameters for modulation. 
These parameters are then used to modulate the adaptive layer normalization (AdaLN)~\cite{dit} followed by a self- or cross-attention:
\begin{align}
    h' & = \text{Attention}(h \odot (\gamma + 1) + \beta),
\end{align}
where $h$ is the mask feature before modulation. 
The resulting feature is then further modulated via the gating mechanism~\cite{dit} with a residual skip connection:
\begin{align}
    \hat{h} & = h + \alpha \odot h_{\text{out}}.
\end{align}

\begin{table}[t]
    \centering
    \small
    \caption{The IoU scores of different degradation schedules.}
    \label{tab:schedules}
    \begin{tabular}{lcccc}
        \toprule
        Schedule & Linear & Constant & Instant \\
        \midrule
        IoU (\%) $\uparrow$ & 71.87 & 68.01 & \textbf{76.23} \\
        \bottomrule
    \end{tabular}
    \vspace{-4mm}
\end{table}

\begin{figure*}[t]
  \centering
  \hfill
  \includegraphics[width=\linewidth]{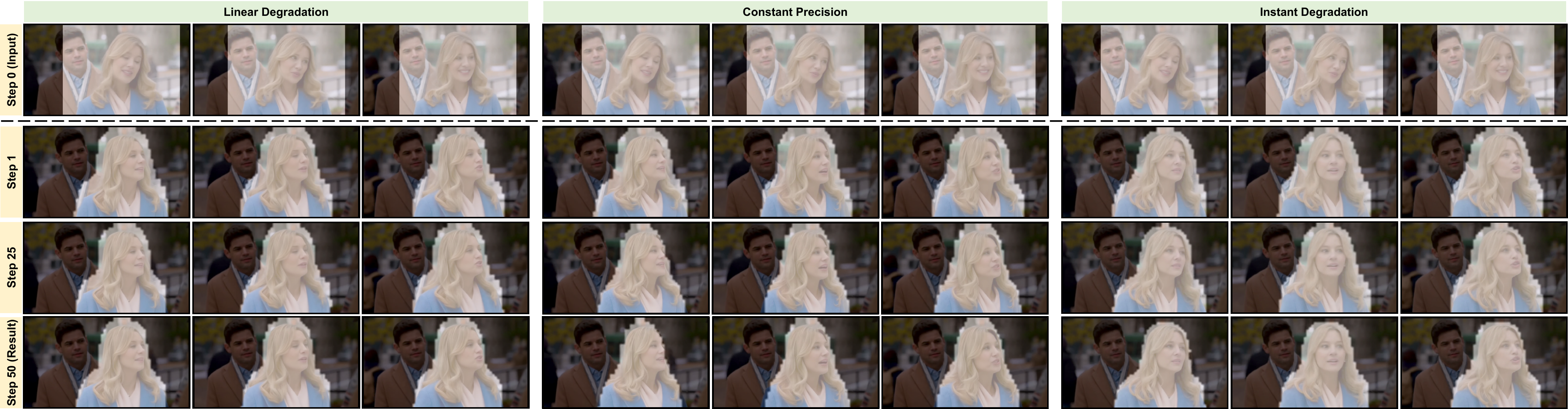}
  \caption{Visual comparison of the mask refinement process under different degradation schedules. \textbf{Top:} Original videos with coarse masks. \textbf{Rows 2-4:} Editing results with evolving refined masks.}
  \vspace{-3mm}
  \label{fig:mask_refine}
  \vspace{-1mm}
\end{figure*}

\subsection{Iterative Mask Refinement Details}

As illustrated in Sec.~\textcolor{cvprblue}{4.2}, the Granularity-Aware Mask Refiner (GAMR) is designed to iteratively refine the instance mask throughout the ODE solving process, guided by the precision factor $p$ and a predefined degradation schedule.  We investigate three distinct schedules:
\textit{(i) Linear degradation.} The precision factor $p$ decreases linearly from its initial value down to $0$ throughout the inference steps. 
\textit{(ii) Constant degradation.} The precision factor $p$ remains fixed at its initial value throughout the inference process. 
\textit{(iii) Instant degradation.} The precision factor $p$ is set to its initial value only for the first step (\ie, $k=0$). For all subsequent steps (\ie, $k \ge 1$), the factor is instantly set to $p=1$, allowing only subtle mask refinement.

As shown in \cref{fig:mask_refine}, the linear and constant schedules are generally suboptimal, because the mask is largely optimized during the initial step ($k=0$). If $p$ decays too slowly, the AMR is guided by an overly-coarse mask granularity. Consequently, the model struggles in predicting the target boundary, causing unnecessary shrinkage in the refined mask and leading to unnatural visual content (\eg, the woman’s face is rendered in profile to fit the overly small region). Therefore, the best schedule strategy is the simple yet effective instant degradation, where the first step quickly converges to the target region, while subsequent steps precisely refine the mask.

To validate this conclusion, we randomly sample 40 video clips from the test split of {\sc AVISet} and calculate the Intersection over Union (IoU)~\cite{iou} of the predicted masks. As presented in \cref{tab:schedules}, the instant degradation strategy achieves the best score, demonstrating that its predicted masks best match the ground truth.

\begin{figure*}[t!]
  \centering
  \hfill
  \includegraphics[width=\linewidth]{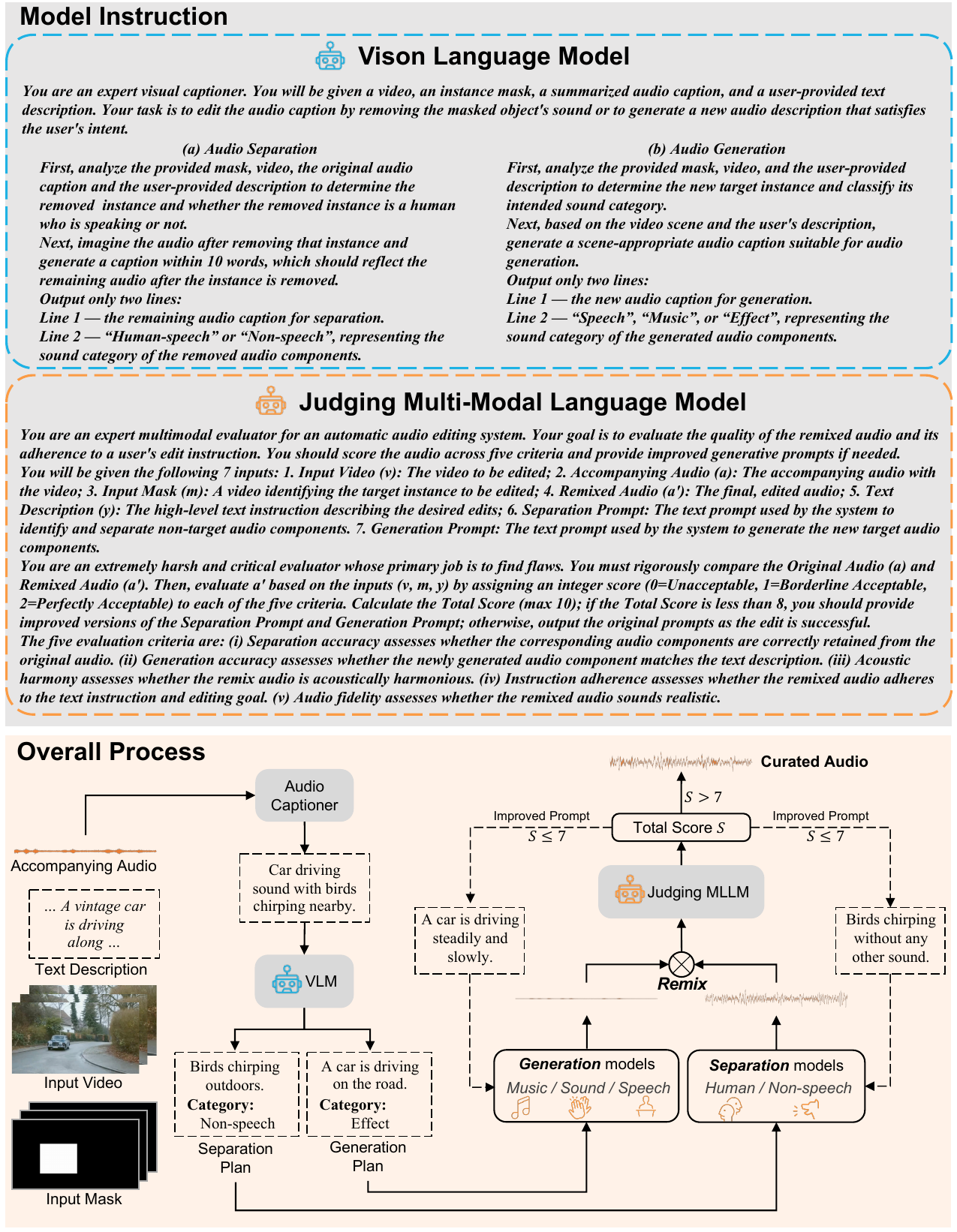}
  \caption{A representative example of the self-feedback audio agent.}
  \vspace{-2mm}
  \label{fig:audioagent}
  \vspace{-2mm}
\end{figure*}

\subsection{Self-Feedback Audio Agent Details} 
As introduced in Sec.~\textcolor{cvprblue}{4.3}, we design a dedicated audio agent to curate the accompanying audio. Since human speech separation is a mature technique that typically requires no text prompts, our prompt refinement loop primarily targets the more challenging non-speech separation and audio generation tasks. 
For clarity, we present a representative execution example in~\cref{fig:audioagent}, illustrating the separate-generate-remix-rework pipeline, including plan initialization, separation, generation, remixing, and quality verification.

\noindent \textbf{Judging MLLM.}
We adopt a judging MLLM to perform quality verification and evaluate the mixed audio across five dimensions: \textit{(i) Separation accuracy} assesses whether the corresponding audio components are correctly retained from the original audio; \textit{(ii) Generation accuracy} assesses whether the newly generated audio component matches the text description; \textit{(iii) Acoustic harmony} assesses whether the remixed audio is acoustically harmonious; \textit{(iv) Instruction adherence} assesses whether the remixed audio adheres to the text instruction and editing goal; and
\textit{(v) Audio fidelity} assesses whether the remixed audio sounds realistic.
We prompt the MLLM to score each dimension on a 3-point scale (\ie, 0: Unacceptable, 1: Borderline acceptable, 2: Perfectly acceptable). Only remixed audios with a total quality score exceeding a predefined threshold $\tau = 7$ are accepted. Otherwise, the audio is rejected, and the MLLM generates an improvement instruction to rework with an iterative refinement loop.

\noindent \textbf{Remix.}
We implement remix as an additive operator without optimization that composites the separation and generation audio components, as a \textit{prerequisite} for final results.

\noindent \textbf{Efficiency.}
We quantitatively assess the operational efficiency of our self-feedback audio agent on a subset of 200 randomly sampled video clips from the test set. On average, the agent requires only 1.67 rework iterations per clip to achieve satisfactory results, with a total processing time averaging 69.9 seconds. Specifically, the initial editing plan generation requires 27.3 seconds, leaving 42.6 seconds dedicated to the iterative self-feedback loop.

\subsection{Inference Costs}
\label{sec:supp_efficiency}

We evaluate the inference time and peak VRAM of our framework and the comparison methods, averaged over 100 clips using a single NVIDIA A100 GPU. As shown in \cref{tab:inference_cost}, AVI-Edit operates over 3$\times$ faster than both AvED~\cite{aved} and VACE-Foley~\cite{vace, hunyuan-foley}, while requiring significantly less peak VRAM than Ovi~\cite{ovi}.

\begin{table}[t]
    \centering
    \small
    \caption{The inference time and peak VRAM of comparison with state-of-the-art methods.}
    \vspace{-2mm}
    \label{tab:inference_cost}
    \resizebox{\linewidth}{!}{
    \begin{tabular}{lcccc}
        \toprule
        Method & AvED & Ovi & VACE-Foley & AVI-Edit \\
        \midrule
        Time (s) $\downarrow$ & 1140.5 & \textbf{232.6} & 1003.6 & 311.3 \\
        VRAM (GB) $\downarrow$ & 37.6 & 59.6 & \textbf{33.2} & 36.2 \\
        \bottomrule
    \end{tabular}
    }
    \vspace{-4mm}
\end{table}

\subsection{Evaluation Metrics Details} 
As described in Sec.~\textcolor{cvprblue}{5.1}, we adopt seven metrics to quantitatively evaluate performance and present the results in Tab.~\textcolor{cvprblue}{1} of the main paper. For clarification, we detail these metrics as follows:

\begin{itemize}

    \item \textbf{FVD} assesses the video quality, calculating the Frechét video distance~\cite{fvd} between real and generated video features, extracted by the Inception Network.

    \item \textbf{IS} assesses the fidelity and diversity of generated frames, calculating the Inception score~\cite{is} using the softmax outputs of the Inception classifier.

    \item \textbf{FC} assesses the temporal consistency across video frames, calculating the cosine similarity between the embeddings of consecutive video frames, extracted by the CLIP visual encoder~\cite{clip}.
        
    \item \textbf{TC} assesses the text-video consistency, calculating the cosine similarity between the generated video and text embeddings extracted by the VideoCLIP-XL~\cite{videoclipxl}.

    \item \textbf{AC} assesses the audio-video consistency, calculating the cosine similarity between the generated audio and video embeddings extracted by the ImageBind~\cite{imagebind}.

    \item \textbf{Sync-C/D} are metrics that assess the audio-lip synchronization. They first calculate the Euclidean distance between audio embeddings and mouth-region video embeddings, extracted by the SyncNet~\cite{sync}. Next, Sync-C is defined as the difference between the minimum and the median of these Euclidean distances, and Sync-D is defined as the minimum of all computed Euclidean distances.

\end{itemize}

\subsection{Organization of Supplementary Video}
We provide a supplementary video to dynamically showcase our audio-sync instance editing results. The video is structured as follows:
\textit{(i)} \textbf{Representative application scenarios:} We demonstrate four representative editing scenarios to highlight the diverse and compelling applications of our framework (Fig.~\textcolor{cvprblue}{1}).
\textit{(ii)} \textbf{Additional application scenarios:} We showcase four extended application scenarios to further demonstrate the versatility of our framework (\cref{fig:supp_application}).
\textit{(iii)} \textbf{Optional control contexts:} We visualize the integration of optional control contexts (\ie, scribble, pose, and reference image) to facilitate fine-grained editing (Fig.~\textcolor{cvprblue}{5}).
\textit{(iv)} \textbf{Comparison and ablation study:} Finally, we provide audio-sync instance editing comparisons against state-of-the-art methods (Fig.~\textcolor{cvprblue}{3}), followed by visual ablations to verify the effectiveness of each component (Fig.~\textcolor{cvprblue}{4}).

%% file: main.bbl
\begin{thebibliography}{74}
\providecommand{\natexlab}[1]{#1}
\providecommand{\url}[1]{\texttt{#1}}
\expandafter\ifx\csname urlstyle\endcsname\relax
  \providecommand{\doi}[1]{doi: #1}\else
  \providecommand{\doi}{doi: \begingroup \urlstyle{rm}\Url}\fi

\bibitem[ele()]{elevenlabs}
Eleven{L}abs.
\newblock \url{https://elevenlabs.io}.

\bibitem[scr()]{scribe}
{ElevenLabs} {Scribe}.
\newblock \url{https://elevenlabs.io/blog/meet-scribe}.

\bibitem[sor()]{sora2}
Sora 2.
\newblock \url{https://openai.com/index/sora-2/}.

\bibitem[veo()]{veo3}
Veo 3.
\newblock \url{https://aistudio.google.com/models/veo-3}.

\bibitem[Bai et~al.(2025)Bai, Chen, Liu, Wang, Ge, Song, Dang, Wang, Wang, Tang, et~al.]{qwen-vl}
Shuai Bai, Keqin Chen, Xuejing Liu, Jialin Wang, Wenbin Ge, Sibo Song, Kai Dang, Peng Wang, Shijie Wang, Jun Tang, et~al.
\newblock Qwen2. 5-vl technical report.
\newblock \emph{arXiv:2502.13923}, 2025.

\bibitem[Bain et~al.(2020)Bain, Nagrani, Brown, and Zisserman]{cdm}
Max Bain, Arsha Nagrani, Andrew Brown, and Andrew Zisserman.
\newblock Condensed movies: Story based retrieval with contextual embeddings.
\newblock In \emph{ACCV}, 2020.

\bibitem[Bara et~al.(2022)Bara, Pokrovskaia, Ababkova, Brusakova, and Korban]{bara2022artificial}
Bibars Al~Haj Bara, Nadezhda~N Pokrovskaia, Marianna~Yu Ababkova, Irina~A Brusakova, and Anastasia~A Korban.
\newblock Artificial intelligence for advertising and media: machine learning and neural networks.
\newblock In \emph{ElConRus}, 2022.

\bibitem[Beliaev and Ginsburg(2021)]{talknet}
Stanislav Beliaev and Boris Ginsburg.
\newblock Talk{N}et 2: Non-autoregressive depth-wise separable convolutional model for speech synthesis with explicit pitch and duration prediction.
\newblock \emph{arXiv:2104.08189}, 2021.

\bibitem[Bian et~al.(2025)Bian, Zhang, Ju, Cao, Xie, Shan, and Xu]{videopainter}
Yuxuan Bian, Zhaoyang Zhang, Xuan Ju, Mingdeng Cao, Liangbin Xie, Ying Shan, and Qiang Xu.
\newblock Video{P}ainter: Any-length video inpainting and editing with plug-and-play context control.
\newblock In \emph{SIGGRAPH}, 2025.

\bibitem[Blattmann et~al.(2023)Blattmann, Dockhorn, Kulal, Mendelevitch, Kilian, Lorenz, Levi, English, Voleti, Letts, et~al.]{svd}
Andreas Blattmann, Tim Dockhorn, Sumith Kulal, Daniel Mendelevitch, Maciej Kilian, Dominik Lorenz, Yam Levi, Zion English, Vikram Voleti, Adam Letts, et~al.
\newblock Stable video diffusion: Scaling latent video diffusion models to large datasets.
\newblock \emph{arXiv:2311.15127}, 2023.

\bibitem[Bruce et~al.(2024)Bruce, Dennis, Edwards, Parker-Holder, Shi, Hughes, Lai, Mavalankar, Steigerwald, Apps, et~al.]{genie}
Jake Bruce, Michael~D Dennis, Ashley Edwards, Jack Parker-Holder, Yuge Shi, Edward Hughes, Matthew Lai, Aditi Mavalankar, Richie Steigerwald, Chris Apps, et~al.
\newblock Genie: Generative interactive environments.
\newblock In \emph{Forty-first International Conference on Machine Learning}, 2024.

\bibitem[Castellano(2025)]{pyscenedetect}
Brandon Castellano.
\newblock {PySceneDetect}, 2025.
\newblock \url{https://github.com/Breakthrough/PySceneDetect}.

\bibitem[Chen et~al.(2020)Chen, Xie, Vedaldi, and Zisserman]{vggsound}
Honglie Chen, Weidi Xie, Andrea Vedaldi, and Andrew Zisserman.
\newblock {VGGS}ound: A large-scale audio-visual dataset.
\newblock In \emph{ICASSP}, 2020.

\bibitem[Chen et~al.(2025{\natexlab{a}})Chen, Cui, Zhang, Zhang, Zhou, Li, Tang, Liu, Liao, Chen, et~al.]{midas}
Ming Chen, Liyuan Cui, Wenyuan Zhang, Haoxian Zhang, Yan Zhou, Xiaohan Li, Songlin Tang, Jiwen Liu, Borui Liao, Hejia Chen, et~al.
\newblock {MIDAS}: Multimodal interactive digital-human synthesis via real-time autoregressive video generation.
\newblock \emph{arXiv:2508.19320}, 2025{\natexlab{a}}.

\bibitem[Chen et~al.(2025{\natexlab{b}})Chen, He, Ma, and Ma]{contextflow}
Yiyang Chen, Xuanhua He, Xiujun Ma, and Yue Ma.
\newblock Context{F}low: Training-free video object editing via adaptive context enrichment.
\newblock \emph{arXiv:2509.17818}, 2025{\natexlab{b}}.

\bibitem[Chen et~al.(2025{\natexlab{c}})Chen, Wang, Liu, Chu, Zhang, Tian, and Yang]{odiscoedit}
Yuqing Chen, Junjie Wang, Lin Liu, Ruihang Chu, Xiaopeng Zhang, Qi Tian, and Yujiu Yang.
\newblock O-{D}is{C}o-{E}dit: Object distortion control for unified realistic video editing.
\newblock \emph{arXiv:2509.01596}, 2025{\natexlab{c}}.

\bibitem[Cheng et~al.(2024)Cheng, Zheng, Wang, Fang, Zhang, Huang, Ma, Ji, Zuo, Jin, et~al.]{omnisep}
Xize Cheng, Siqi Zheng, Zehan Wang, Minghui Fang, Ziang Zhang, Rongjie Huang, Ziyang Ma, Shengpeng Ji, Jialong Zuo, Tao Jin, et~al.
\newblock Omni{S}ep: Unified omni-modality sound separation with query-mixup.
\newblock \emph{arXiv:2410.21269}, 2024.

\bibitem[Chung and Zisserman(2017)]{sync}
Joon~Son Chung and Andrew Zisserman.
\newblock Out of time: automated lip sync in the wild.
\newblock In \emph{ACCV Workshops}, 2017.

\bibitem[Esser et~al.(2021)Esser, Rombach, and Ommer]{vqgan}
Patrick Esser, Robin Rombach, and Björn Ommer.
\newblock Taming transformers for high-resolution image synthesis.
\newblock In \emph{CVPR}, 2021.

\bibitem[Everingham et~al.(2010)Everingham, Van~Gool, Williams, Winn, and Zisserman]{iou}
Mark Everingham, Luc Van~Gool, Christopher~KI Williams, John Winn, and Andrew Zisserman.
\newblock The pascal visual object classes (voc) challenge.
\newblock \emph{IJCV}, 2010.

\bibitem[Fu et~al.(2025)Fu, Si, Wang, Zhou, Sun, Luo, Hu, Zhang, and Li]{objavedit}
Youquan Fu, Ruiyang Si, Hongfa Wang, Dongzhan Zhou, Jiacheng Sun, Ping Luo, Di Hu, Hongyuan Zhang, and Xuelong Li.
\newblock Object-{AVE}dit: An object-level audio-visual editing model.
\newblock \emph{arXiv:2510.00050}, 2025.

\bibitem[Ge et~al.(2022)Ge, Hayes, Yang, Yin, Pang, Jacobs, Huang, and Parikh]{tats}
Songwei Ge, Thomas Hayes, Harry Yang, Xi Yin, Guan Pang, David Jacobs, Jia-Bin Huang, and Devi Parikh.
\newblock Long video generation with time-agnostic vqgan and time-sensitive transformer.
\newblock In \emph{ECCV}, 2022.

\bibitem[Gemmeke et~al.(2017)Gemmeke, Ellis, Freedman, Jansen, Lawrence, Moore, Plakal, and Ritter]{audioset}
Jort~F Gemmeke, Daniel~PW Ellis, Dylan Freedman, Aren Jansen, Wade Lawrence, R~Channing Moore, Manoj Plakal, and Marvin Ritter.
\newblock Audio{S}et: An ontology and human-labeled dataset for audio events.
\newblock In \emph{ICASSP}, 2017.

\bibitem[Ghermi et~al.(2024)Ghermi, Wang, Kalogeiton, and Laptev]{sf20k}
Ridouane Ghermi, Xi Wang, Vicky Kalogeiton, and Ivan Laptev.
\newblock Short film dataset {(SFD):} {A} benchmark for story-level video understanding.
\newblock \emph{CoRR}, 2024.

\bibitem[Girdhar et~al.(2023)Girdhar, El-Nouby, Liu, Singh, Alwala, Joulin, and Misra]{imagebind}
Rohit Girdhar, Alaaeldin El-Nouby, Zhuang Liu, Mannat Singh, Kalyan~Vasudev Alwala, Armand Joulin, and Ishan Misra.
\newblock Image{B}ind: One embedding space to bind them all.
\newblock In \emph{CVPR}, 2023.

\bibitem[Guo et~al.(2023)Guo, Yang, Rao, Liang, Wang, Qiao, Agrawala, Lin, and Dai]{animatediff}
Yuwei Guo, Ceyuan Yang, Anyi Rao, Zhengyang Liang, Yaohui Wang, Yu Qiao, Maneesh Agrawala, Dahua Lin, and Bo Dai.
\newblock Animate{D}iff: Animate your personalized text-to-image diffusion models without specific tuning.
\newblock \emph{arXiv:2307.04725}, 2023.

\bibitem[Hennequin et~al.(2020)Hennequin, Khlif, Voituret, and Moussallam]{Spleeter}
Romain Hennequin, Anis Khlif, Felix Voituret, and Manuel Moussallam.
\newblock Spleeter: a fast and efficient music source separation tool with pre-trained models.
\newblock \emph{Journal of Open Source Software}, 2020.

\bibitem[Jeong et~al.(2021)Jeong, Doh, and Kwon]{Träumerai}
Dasaem Jeong, Seungheon Doh, and Taegyun Kwon.
\newblock Tr{\"a}umerai: Dreaming music with stylegan.
\newblock \emph{arXiv:2102.04680}, 2021.

\bibitem[Jiang et~al.(2025)Jiang, Han, Mao, Zhang, Pan, and Liu]{vace}
Zeyinzi Jiang, Zhen Han, Chaojie Mao, Jingfeng Zhang, Yulin Pan, and Yu Liu.
\newblock {VACE}: All-in-one video creation and editing.
\newblock \emph{arXiv:2503.07598}, 2025.

\bibitem[Kara et~al.(2024)Kara, Kurtkaya, Yesiltepe, Rehg, and Yanardag]{rave}
Ozgur Kara, Bariscan Kurtkaya, Hidir Yesiltepe, James~M Rehg, and Pinar Yanardag.
\newblock {RAVE}: Randomized noise shuffling for fast and consistent video editing with diffusion models.
\newblock In \emph{CVPR}, 2024.

\bibitem[Karras et~al.(2019)Karras, Laine, and Aila]{stylegan}
Tero Karras, Samuli Laine, and Timo Aila.
\newblock A style-based generator architecture for generative adversarial networks.
\newblock In \emph{CVPR}, 2019.

\bibitem[Kingma and Ba(2014)]{adam}
Diederik~P Kingma and Jimmy Ba.
\newblock Adam: A method for stochastic optimization.
\newblock \emph{arXiv:1412.6980}, 2014.

\bibitem[Kong et~al.(2024)Kong, Tian, Zhang, Min, Dai, Zhou, Xiong, Li, Wu, Zhang, et~al.]{hunyuan}
Weijie Kong, Qi Tian, Zijian Zhang, Rox Min, Zuozhuo Dai, Jin Zhou, Jiangfeng Xiong, Xin Li, Bo Wu, Jianwei Zhang, et~al.
\newblock Hunyuan{V}ideo: A systematic framework for large video generative models.
\newblock \emph{arXiv:2412.03603}, 2024.

\bibitem[Labs et~al.(2025)Labs, Batifol, Blattmann, Boesel, Consul, Diagne, Dockhorn, English, English, Esser, et~al.]{flux}
Black~Forest Labs, Stephen Batifol, Andreas Blattmann, Frederic Boesel, Saksham Consul, Cyril Diagne, Tim Dockhorn, Jack English, Zion English, Patrick Esser, et~al.
\newblock {FLUX}. 1 {K}ontext: Flow matching for in-context image generation and editing in latent space.
\newblock \emph{arXiv:2506.15742}, 2025.

\bibitem[Lee et~al.(2020)Lee, Lin, Shih, Kuo, and Su]{sound2sight}
Cheng-Che Lee, Wan-Yi Lin, Yen-Ting Shih, Pei-Yi Kuo, and Li Su.
\newblock Crossing you in style: Cross-modal style transfer from music to visual arts.
\newblock In \emph{ACM MM}, 2020.

\bibitem[Lee et~al.(2022)Lee, Oh, Byeon, Kim, Ryoo, Yoon, Cho, Bae, Kim, and Kim]{lee2022sound}
Seung~Hyun Lee, Gyeongrok Oh, Wonmin Byeon, Chanyoung Kim, Won~Jeong Ryoo, Sang~Ho Yoon, Hyunjun Cho, Jihyun Bae, Jinkyu Kim, and Sangpil Kim.
\newblock Sound-guided semantic video generation.
\newblock In \emph{ECCV}, 2022.

\bibitem[Lee et~al.(2023{\natexlab{a}})Lee, Kim, Yoo, Yang, Cho, Kim, Chang, Kim, and Kim]{Soundini}
Seung~Hyun Lee, Sieun Kim, Innfarn Yoo, Feng Yang, Donghyeon Cho, Youngseo Kim, Huiwen Chang, Jinkyu Kim, and Sangpil Kim.
\newblock Soundini: Sound-guided diffusion for natural video editing.
\newblock \emph{arXiv:2304.06818}, 2023{\natexlab{a}}.

\bibitem[Lee et~al.(2023{\natexlab{b}})Lee, Kang, Kim, and Kim]{lee2023generating}
Taegyeong Lee, Jeonghun Kang, Hyeonyu Kim, and Taehwan Kim.
\newblock Generating realistic images from in-the-wild sounds.
\newblock In \emph{ICCV}, 2023{\natexlab{b}}.

\bibitem[Li et~al.(2024)Li, Ma, Yang, and Yang]{vidtome}
Xirui Li, Chao Ma, Xiaokang Yang, and Ming-Hsuan Yang.
\newblock Vid{T}o{M}e: Video token merging for zero-shot video editing.
\newblock In \emph{CVPR}, 2024.

\bibitem[Lin et~al.(2017)Lin, Goyal, Girshick, He, and Doll{\'a}r]{focal_loss}
Tsung-Yi Lin, Priya Goyal, Ross Girshick, Kaiming He, and Piotr Doll{\'a}r.
\newblock Focal loss for dense object detection.
\newblock In \emph{ICCV}, 2017.

\bibitem[Lin et~al.(2025)Lin, Lin, Yang, Li, Wang, Lin, Wang, Bertasius, and Wang]{aved}
Yan-Bo Lin, Kevin Lin, Zhengyuan Yang, Linjie Li, Jianfeng Wang, Chung-Ching Lin, Xiaofei Wang, Gedas Bertasius, and Lijuan Wang.
\newblock Zero-shot audio-visual editing via cross-modal delta denoising.
\newblock \emph{arXiv:2503.20782}, 2025.

\bibitem[Lipman et~al.(2022)Lipman, Chen, Ben-Hamu, Nickel, and Le]{flow_matching}
Yaron Lipman, Ricky~TQ Chen, Heli Ben-Hamu, Maximilian Nickel, and Matt Le.
\newblock Flow matching for generative modeling.
\newblock \emph{arXiv:2210.02747}, 2022.

\bibitem[Low et~al.(2025)Low, Wang, and Katyal]{ovi}
Chetwin Low, Weimin Wang, and Calder Katyal.
\newblock Ovi: Twin backbone cross-modal fusion for audio-video generation.
\newblock \emph{arXiv:2510.01284}, 2025.

\bibitem[Manor and Michaeli(2024)]{audioeditcode}
Hila Manor and Tomer Michaeli.
\newblock Zero-shot unsupervised and text-based audio editing using ddpm inversion.
\newblock In \emph{ICML}, 2024.

\bibitem[Oh et~al.(2019)Oh, Dekel, Kim, Mosseri, Freeman, Rubinstein, and Matusik]{speech2face}
Tae-Hyun Oh, Tali Dekel, Changil Kim, Inbar Mosseri, William~T Freeman, Michael Rubinstein, and Wojciech Matusik.
\newblock Speech2{F}ace: Learning the face behind a voice.
\newblock In \emph{CVPR}, 2019.

\bibitem[Peebles and Xie(2023)]{dit}
William Peebles and Saining Xie.
\newblock Scalable diffusion models with transformers.
\newblock In \emph{ICCV}, 2023.

\bibitem[Qi et~al.(2023)Qi, Cun, Zhang, Lei, Wang, Shan, and Chen]{fatazero}
Chenyang Qi, Xiaodong Cun, Yong Zhang, Chenyang Lei, Xintao Wang, Ying Shan, and Qifeng Chen.
\newblock Fate{Z}ero: Fusing attentions for zero-shot text-based video editing.
\newblock In \emph{ICCV}, 2023.

\bibitem[Radford et~al.(2021)Radford, Kim, Hallacy, Ramesh, Goh, Agarwal, Sastry, Askell, Mishkin, Clark, et~al.]{clip}
Alec Radford, Jong~Wook Kim, Chris Hallacy, Aditya Ramesh, Gabriel Goh, Sandhini Agarwal, Girish Sastry, Amanda Askell, Pamela Mishkin, Jack Clark, et~al.
\newblock Learning transferable visual models from natural language supervision.
\newblock In \emph{ICML}, 2021.

\bibitem[Ren et~al.(2024)Ren, Liu, Zeng, Lin, Li, Cao, Chen, Huang, Chen, Yan, et~al.]{grounded-sam2}
Tianhe Ren, Shilong Liu, Ailing Zeng, Jing Lin, Kunchang Li, He Cao, Jiayu Chen, Xinyu Huang, Yukang Chen, Feng Yan, et~al.
\newblock Grounded {SAM}: Assembling open-world models for diverse visual tasks.
\newblock \emph{arXiv:2401.14159}, 2024.

\bibitem[Robert(2025)]{pydub}
James Robert.
\newblock Pydub, 2025.
\newblock \url{https://github.com/jiaaro/pydub}.

\bibitem[Rombach et~al.(2022)Rombach, Blattmann, Lorenz, Esser, and Ommer]{sd}
Robin Rombach, Andreas Blattmann, Dominik Lorenz, Patrick Esser, and Bj{\"o}rn Ommer.
\newblock High-resolution image synthesis with latent diffusion models.
\newblock In \emph{CVPR}, 2022.

\bibitem[Salimans et~al.(2016)Salimans, Goodfellow, Zaremba, Cheung, Radford, and Chen]{is}
Tim Salimans, Ian Goodfellow, Wojciech Zaremba, Vicki Cheung, Alec Radford, and Xi Chen.
\newblock Improved techniques for training gans.
\newblock In \emph{NIPS}, 2016.

\bibitem[Schuhmann(2025)]{ase}
Christoph Schuhmann.
\newblock Improved aesthetic predictor, 2025.
\newblock \url{https://github.com/christophschuhmann/improved-aesthetic-predictor}.

\bibitem[Shan et~al.(2025)Shan, Li, Cui, Yang, Wang, Yang, Zhou, and Zhong]{hunyuan-foley}
Sizhe Shan, Qiulin Li, Yutao Cui, Miles Yang, Yuehai Wang, Qun Yang, Jin Zhou, and Zhao Zhong.
\newblock Hunyuan{V}ideo-{F}oley: Multimodal diffusion with representation alignment for high-fidelity foley audio generation.
\newblock \emph{arXiv:2508.16930}, 2025.

\bibitem[Shen et~al.(2024)Shen, Quan, Zhu, Xiao, and Yang]{audioscenic}
Kaixin Shen, Ruijie Quan, Linchao Zhu, Jun Xiao, and Yi Yang.
\newblock Audio{S}cenic: Audio-driven video scene editing.
\newblock \emph{arXiv:2404.16581}, 2024.

\bibitem[Teed and Deng(2020)]{raft}
Zachary Teed and Jia Deng.
\newblock Raft: Recurrent all-pairs field transforms for optical flow.
\newblock In \emph{ECCV}, 2020.

\bibitem[Tjandra et~al.(2025)Tjandra, Wu, Guo, Hoffman, Ellis, Vyas, Shi, Chen, Le, Zacharov, et~al.]{audiobox-aesthetics}
Andros Tjandra, Yi-Chiao Wu, Baishan Guo, John Hoffman, Brian Ellis, Apoorv Vyas, Bowen Shi, Sanyuan Chen, Matt Le, Nick Zacharov, et~al.
\newblock Meta audiobox aesthetics: Unified automatic quality assessment for speech, music, and sound.
\newblock \emph{arXiv:2502.05139}, 2025.

\bibitem[Uhlich et~al.(2024)Uhlich, Fabbro, Hirano, Takahashi, Wichern, Roux, Chakraborty, Mohanty, Li, Luo, Yu, Gu, Solovyev, Stempkovskiy, Habruseva, Sukhovei, and Mitsufuji]{cdx}
Stefan Uhlich, Giorgio Fabbro, Masato Hirano, Shusuke Takahashi, Gordon Wichern, Jonathan~Le Roux, Dipam Chakraborty, Sharada Mohanty, Kai Li, Yi Luo, Jianwei Yu, Rongzhi Gu, Roman~A. Solovyev, Alexander~L. Stempkovskiy, Tatiana Habruseva, Mikhail Sukhovei, and Yuki Mitsufuji.
\newblock The sound demixing challenge 2023 - cinematic demixing track.
\newblock \emph{Trans. Int. Soc. Music. Inf. Retr.}, 2024.

\bibitem[Unterthiner et~al.(2018)Unterthiner, Van~Steenkiste, Kurach, Marinier, Michalski, and Gelly]{fvd}
Thomas Unterthiner, Sjoerd Van~Steenkiste, Karol Kurach, Raphael Marinier, Marcin Michalski, and Sylvain Gelly.
\newblock Towards accurate generative models of video: A new metric \& challenges.
\newblock \emph{arXiv:1812.01717}, 2018.

\bibitem[Wan et~al.(2025)Wan, Wang, Ai, Wen, Mao, Xie, Chen, Yu, Zhao, Yang, et~al.]{wan}
Team Wan, Ang Wang, Baole Ai, Bin Wen, Chaojie Mao, Chen-Wei Xie, Di Chen, Feiwu Yu, Haiming Zhao, Jianxiao Yang, et~al.
\newblock Wan: Open and advanced large-scale video generative models.
\newblock \emph{arXiv:2503.20314}, 2025.

\bibitem[Wang et~al.(2025{\natexlab{a}})Wang, Zuo, Li, Chen, Liao, Zhou, Yin, Dai, Jiang, and Yu]{uniVerse-1}
Duomin Wang, Wei Zuo, Aojie Li, Ling-Hao Chen, Xinyao Liao, Deyu Zhou, Zixin Yin, Xili Dai, Daxin Jiang, and Gang Yu.
\newblock Uni{V}erse-1: Unified audio-video generation via stitching of experts.
\newblock \emph{arXiv:2509.06155}, 2025{\natexlab{a}}.

\bibitem[Wang et~al.(2024)Wang, Wang, Huang, Huang, and Jin]{videoclipxl}
Jiapeng Wang, Chengyu Wang, Kunzhe Huang, Jun Huang, and Lianwen Jin.
\newblock Video{CLIP}-{XL}: Advancing long description understanding for video clip models.
\newblock \emph{arXiv:2410.00741}, 2024.

\bibitem[Wang et~al.(2025{\natexlab{b}})Wang, Yang, Jiang, Liang, Lin, Zheng, Yang, and Lin]{interacthuman}
Zhenzhi Wang, Jiaqi Yang, Jianwen Jiang, Chao Liang, Gaojie Lin, Zerong Zheng, Ceyuan Yang, and Dahua Lin.
\newblock Inter{A}ct{H}uman: Multi-concept human animation with layout-aligned audio conditions.
\newblock \emph{arXiv:2506.09984}, 2025{\natexlab{b}}.

\bibitem[Weng et~al.(2025{\natexlab{a}})Weng, Zheng, Chang, Li, Shi, and Wang]{mtv}
Shuchen Weng, Haojie Zheng, Zheng Chang, Si Li, Boxin Shi, and Xinlong Wang.
\newblock Audio-sync video generation with multi-stream temporal control.
\newblock In \emph{NIPS}, 2025{\natexlab{a}}.

\bibitem[Weng et~al.(2025{\natexlab{b}})Weng, Zheng, Zhang, Hong, Jiang, Li, and Shi]{vires}
Shuchen Weng, Haojie Zheng, Peixuan Zhang, Yuchen Hong, Han Jiang, Si Li, and Boxin Shi.
\newblock V{IRES}: Video instance repainting via sketch and text guided generation.
\newblock In \emph{{CVPR}}, 2025{\natexlab{b}}.

\bibitem[Wu et~al.(2023)Wu, Ge, Wang, Lei, Gu, Shi, Hsu, Shan, Qie, and Shou]{tune-a-video}
Jay~Zhangjie Wu, Yixiao Ge, Xintao Wang, Stan~Weixian Lei, Yuchao Gu, Yufei Shi, Wynne Hsu, Ying Shan, Xiaohu Qie, and Mike~Zheng Shou.
\newblock Tune-{A}-{V}ideo: One-shot tuning of image diffusion models for text-to-video generation.
\newblock In \emph{ICCV}, 2023.

\bibitem[Wu et~al.(2024)Wu, Liu, Zhu, Xia, Feng, Wang, Lin, Shen, and Shou]{moviebench}
Weijia Wu, Mingyu Liu, Zeyu Zhu, Xi Xia, Haoen Feng, Wen Wang, Kevin~Qinghong Lin, Chunhua Shen, and Mike~Zheng Shou.
\newblock Movie{B}ench: A hierarchical movie level dataset for long video generation.
\newblock \emph{arXiv:2411.15262}, 2024.

\bibitem[Xu et~al.(2025)Xu, Guo, Hu, Chu, Wang, He, Wang, Shi, He, Zhu, et~al.]{qwen-omni}
Jin Xu, Zhifang Guo, Hangrui Hu, Yunfei Chu, Xiong Wang, Jinzheng He, Yuxuan Wang, Xian Shi, Ting He, Xinfa Zhu, et~al.
\newblock Qwen3-omni technical report.
\newblock \emph{arXiv:2509.17765}, 2025.

\bibitem[Yang et~al.(2025{\natexlab{a}})Yang, Li, Yang, Zhang, Hui, Zheng, Yu, Gao, Huang, Lv, et~al.]{qwen}
An Yang, Anfeng Li, Baosong Yang, Beichen Zhang, Binyuan Hui, Bo Zheng, Bowen Yu, Chang Gao, Chengen Huang, Chenxu Lv, et~al.
\newblock Qwen3 technical report.
\newblock \emph{arXiv:2505.09388}, 2025{\natexlab{a}}.

\bibitem[Yang et~al.(2025{\natexlab{b}})Yang, Li, Cun, Wang, Li, Shan, and Zhang]{gencompositor}
Shuzhou Yang, Xiaoyu Li, Xiaodong Cun, Guangzhi Wang, Lingen Li, Ying Shan, and Jian Zhang.
\newblock Gen{C}ompositor: generative video compositing with diffusion transformer.
\newblock \emph{arXiv:2509.02460}, 2025{\natexlab{b}}.

\bibitem[Yang et~al.(2025{\natexlab{c}})Yang, Teng, Zheng, Ding, Huang, Xu, Yang, Hong, Zhang, Feng, et~al.]{cogvideox}
Zhuoyi Yang, Jiayan Teng, Wendi Zheng, Ming Ding, Shiyu Huang, Jiazheng Xu, Yuanming Yang, Wenyi Hong, Xiaohan Zhang, Guanyu Feng, et~al.
\newblock Cog{V}ideox: Text-to-video diffusion models with an expert transformer.
\newblock In \emph{ICLR}, 2025{\natexlab{c}}.

\bibitem[Yariv et~al.(2023)Yariv, Gat, Wolf, Adi, and Schwartz]{audiotoken}
Guy Yariv, Itai Gat, Lior Wolf, Yossi Adi, and Idan Schwartz.
\newblock Audio{T}oken: Adaptation of text-conditioned diffusion models for audio-to-image generation.
\newblock \emph{arXiv:2305.13050}, 2023.

\bibitem[Zhang et~al.(2025{\natexlab{a}})Zhang, Jia, Liu, Weng, Li, and Shi]{zhang2025stage}
Peixuan Zhang, Zijian Jia, Kaiqi Liu, Shuchen Weng, Si Li, and Boxin Shi.
\newblock Stage: Storyboard-anchored generation for cinematic multi-shot narrative.
\newblock \emph{arXiv preprint arXiv:2512.12372}, 2025{\natexlab{a}}.

\bibitem[Zhang et~al.(2025{\natexlab{b}})Zhang, Yu, Min, Xin, Wei, Shi, Huang, Kong, Xin, Jiang, Bahuguna, Chan, Hora, Yang, Liang, Bian, Liu, Valencia, Tredinick, Kozlov, Jiang, Huang, Chen, Liu, and Rao]{zhang2025generative}
Ruihan Zhang, Borou Yu, Jiajian Min, Yetong Xin, Zheng Wei, Juncheng~Nemo Shi, Mingzhen Huang, Xianghao Kong, Nix~Liu Xin, Shanshan Jiang, Praagya Bahuguna, Mark Chan, Khushi Hora, Lijian Yang, Yongqi Liang, Runhe Bian, Yunlei Liu, Isabela~Campillo Valencia, Patricia~Morales Tredinick, Ilia Kozlov, Sijia Jiang, Peiwen Huang, Na Chen, Xuanxuan Liu, and Anyi Rao.
\newblock Generative {AI} for film creation: {A} survey of recent advances.
\newblock In \emph{CVPR Workshops}, 2025{\natexlab{b}}.

\end{thebibliography}
